\documentclass[journal]{IEEEtran}
\usepackage{amsmath,amsfonts}
\usepackage{algorithmic}
\usepackage{algorithm}
\usepackage{array}
\usepackage[caption=false,font=normalsize,labelfont=sf,textfont=sf]{subfig}
\usepackage{textcomp}
\usepackage{stfloats}
\usepackage{url}
\usepackage{verbatim}
\usepackage{graphicx}
\usepackage{cite}
\usepackage{hyperref}
\usepackage[T1]{fontenc}
\usepackage{bm}

\usepackage{tikz}
\usetikzlibrary{positioning}
\usepackage{listing}
\usepackage{booktabs}
 \usepackage{ifthen}
 \newboolean{final}
 \setboolean{final}{true}
\newboolean{doubleblind}
\setboolean{doubleblind}{false}

\ifthenelse{\boolean{doubleblind}}{
 
 \newcommand{\anonymize}[1]{\textcolor{blue}{text removed for peer review}}
}{
 
 \newcommand{\anonymize}[1]{#1}
}

\ifthenelse{\boolean{final}}{
	\newcommand{\helppls}[1]{}
	\newcommand{\todo}[1]{}
	
	\newcommand{\hint}[1]{}
	\newcommand{\review}[1]{}
	\newcommand{\optional}[1]{}
	\newcommand{\questionForReview}[1]{}
	\newcommand{\saveSpaceHere}[1]{}

}{
	\newcommand{\todo}[1]{{\color{red}\textbf{[#1]}}}
	
	\newcommand{\hint}[1]{{\color{hint}\textit{[#1]}}}
	\newcommand{\helppls}[1]{{\textcolor{purple}{[Help needed: #1]}}}
	\newcommand{\review}[1]{\textcolor{orange}{[Review required: #1]}}
	\newcommand{\optional}[1]{\textcolor{blue}{[Possibly relevant content: #1]}}
	\newcommand{\questionForReview}[1]{\review{Question for Review: #1}}
	\newcommand{\saveSpaceHere}[1]{\hint{Save space here: #1}}

}

\newcommand{\ctod}{\texttt{c2d}}
\newcommand{\dsharp}{\texttt{dSharp}}
\newcommand{\dfour}{\texttt{d4}}

\newcommand{\ddknnife}{\texttt{ddnnife}}

\newcommand{\ssat}{\texttt{\#}SAT}

\definecolor{blue1}{RGB}{0,105,180} 
\definecolor{blue2}{RGB}{40,87,121}
\definecolor{blue3}{RGB}{0,57,97}
\definecolor{blue4}{RGB}{76,166,230}
\definecolor{blue5}{RGB}{136,191,230}
\definecolor{orange1}{RGB}{255,144,0} 
\definecolor{orange2}{RGB}{171,121,56}
\definecolor{orange3}{RGB}{137,78,0}
\definecolor{orange4}{RGB}{255,181,84}
\definecolor{orange5}{RGB}{255,210,151}
\definecolor{green1}{RGB}{11,215,0} 

\definecolor{green2}{RGB}{52,144,48}
\definecolor{green3}{RGB}{6,116,0}
\definecolor{green4}{RGB}{88,241,80}
\definecolor{green5}{RGB}{148,241,143}
\definecolor{red1}{RGB}{253,0,6} 
\definecolor{red2}{RGB}{170,56,59}
\definecolor{red3}{RGB}{136,0,3}
\definecolor{red4}{RGB}{254,84,88}
\definecolor{red5}{RGB}{254,151,154}

\definecolor{background}{named}{white}
\definecolor{bgborder}{named}{black}
\definecolor{comment}{named}{red3}
\definecolor{hint}{named}{blue3}

\definecolor{blue}{named}{blue1}
\definecolor{green}{named}{green1}
\definecolor{red}{named}{red1}
\definecolor{orange}{named}{orange1}

\definecolor{pdflinkcolor}{named}{blue3}
\definecolor{pdfcitecolor}{named}{green3}

\definecolor{tubsred}{RGB}{130, 30, 60}
\definecolor{eclipseBlue}{RGB}{0, 7, 196}
\definecolor{eclipsePurple}{RGB}{128, 0, 82}

\definecolor{tolbrightred}{HTML}{ee6677}
\definecolor{tolbrightgreen}{HTML}{228833}
\definecolor{tolbrightblue}{HTML}{4477aa}
\definecolor{tolbrightyellow}{HTML}{ccbb44}
\definecolor{tolbrightskyblue}{HTML}{66ccee}
\definecolor{tolbrightpurple}{HTML}{aa3377}
\definecolor{tolbrightgray}{HTML}{bbbbbb}

\usepackage{tikz}
\usetikzlibrary{arrows.meta,shapes,positioning}

\tikzset{
  dag/.style={
    nodes={draw, circle, minimum size = .1cm, inner sep=2pt},
    -{Stealth[length=1mm]},
	level 1/.style={sibling distance=20mm,level distance=0.5cm},
	level 2/.style={sibling distance=10mm,level distance=0.5cm},
	level 3/.style={sibling distance=10mm,level distance=0.5cm}
  },
  font={\tiny\selectfont}
}

\usepackage{xpunctuate}

\providecommand{\eg}[0]{e.g\xperiod}
\providecommand{\ie}[0]{i.e\xperiod}
\providecommand{\etal}[0]{et~al\xperiod}
\providecommand{\cf}[0]{cf\xperiod}

\providecommand{\wrt}[0]{w.r.t\xperiod}

\newcommand{\latexutilautorefoverride}[2]{
	\ifcsname #1\endcsname
		\expandafter\renewcommand\csname #1\endcsname{#2}
	\else
		\expandafter\newcommand\csname #1\endcsname{#2}
	\fi	
}

\latexutilautorefoverride{figureautorefname}{Figure}
\latexutilautorefoverride{chapterautorefname}{Chapter}
\latexutilautorefoverride{equationautorefname}{Equation}
\latexutilautorefoverride{sectionautorefname}{Section}
\latexutilautorefoverride{subsectionautorefname}{Section}
\latexutilautorefoverride{subsubsectionautorefname}{Section}
\latexutilautorefoverride{definitionautorefname}{Definition}
\latexutilautorefoverride{footnoteautorefname}{Footnote}
\latexutilautorefoverride{algorithmautorefname}{Algorithm}

\newboolean{email}
\setboolean{email}{false}
\newboolean{city}
\setboolean{city}{false}

\newcommand{\affchicosundermann}{
	\author{Chico Sundermann}
	\orcid{0000-0002-5239-3307}
	\affiliation{%
		\institution{University of Ulm \\ TU Braunschweig}
		\ifthenelse{\boolean{city}}{\city{Ulm}}{}
		\country{Germany}}
	\ifthenelse{\boolean{email}}{\email{chico.sundermann@uni-ulm.de}}{}
}

\newcommand{\affthomasthuem}{
	\author{Thomas Thüm}
	\orcid{0000-0001-8069-9584}
	\affiliation{%
		\institution{TU Braunschweig}
		\ifthenelse{\boolean{city}}{\city{Braunschweig}}{}
		\country{Germany}}
	\ifthenelse{\boolean{email}}{\email{thomas.thuem@tu-braunschweig.de}}
}

\newcommand{\affheikoraab}{
	\author{Heiko Raab}
	\orcid{0009-0002-2342-9857}
	\affiliation{%
		\institution{University of Ulm}
		\ifthenelse{\boolean{city}}{\city{Ulm}}{}
		\country{Germany}}
	\ifthenelse{\boolean{email}}{\email{heiko.raab@uni-ulm.de}}
}

\newcommand{\affsebastiankrieter}{
	\author{Sebastian Krieter}
	\orcid{0000-0001-7077-7091}
	\affiliation{%
		\institution{TU Braunschweig}
		\ifthenelse{\boolean{city}}{\city{Braunschweig}}{}
		\country{Germany}}
	\ifthenelse{\boolean{email}}{\email{sebastian.krieter@uni-ulm.de}}
}

\newcommand{\afftobiashess}{
	\author{Tobias Heß}
	\orcid{0000-0001-9389-9278}
	\affiliation{%
		\institution{University of Ulm}
		\ifthenelse{\boolean{city}}{\city{Ulm}}{}
		\country{Germany}}
	\ifthenelse{\boolean{email}}{\email{tobias.hess@uni-ulm.de}}
}

\newcommand{\affeliaskuiter}{
	\author{Elias Kuiter}
	\orcid{0000-0003-0429-2461}
	\affiliation{%
		\institution{University of Magdeburg}
		\ifthenelse{\boolean{city}}{\city{Madgeburg}}{}
		\country{Germany}}
		\ifthenelse{\boolean{email}}{\email{kuiter@ovgu.de}}

}

\newcommand{\affinaschaefer}{
	\author{Ina Schaefer}
	\orcid{0000-0002-7153-761X}
	\affiliation{%
		\institution{Karlsruhe Institute of Technology}
		\ifthenelse{\boolean{city}}{\city{Karlsruhe}}{}
		\country{Germany}
	}
	\ifthenelse{\boolean{email}}{\email{ina.schaefer@kit.edu}}
}

\newcommand{\affrahelsundermann}{
	\author{Rahel Sundermann}
	\orcid{0009-0000-3135-2680}
	\affiliation{%
		\institution{University of Ulm}
		\ifthenelse{\boolean{city}}{\city{Ulm}}{}
		\country{Germany}}
	\ifthenelse{\boolean{email}}{\email{rahel.sundermann@uni-ulm.de}}{}
}

\usepackage{pifont}
\usepackage{enumitem}
\usepackage{calc}
\newcommand*{\researchquestion}[1]{\item[\textbf{RQ#1}]}
\newcommand*{\researchquestionlabelwidthof}[1]{\labelsep+\widthof{\textbf{RQ#1}}}

\usepackage{algorithm}
\usepackage{algorithmic}

\usepackage{forest}

\usepackage{nicefrac}

\newcommand{\ptod}{\texttt{p2d}}
\newcommand{\uvltopb}{\texttt{UVL2pb}}

\newcommand{\featurestyle}[1]{\texttt{#1}}

\usepackage{multirow}

\newcommand{\termoppplaceholder}{\ensuremath{\star}}

\newcommand{\researchanswer}[1]{
  \noindent\fbox{\begin{minipage}{0.975\columnwidth}
    #1
    \end{minipage}}
}

\usepackage{forest}
\usepackage{xcolor}
\usetikzlibrary{angles}
\definecolor{drawColor}{RGB}{128 128 128}
\newcommand{\circleSize}{0.2em}
\newcommand{\angleSize}{0.8em}
\forestset{
	/tikz/mandatory/.style={
		circle,fill=drawColor,
		draw=drawColor,
		inner sep=\circleSize
	},
	/tikz/optional/.style={
		circle,
		fill=white,
		draw=drawColor,
		inner sep=\circleSize
	},
	featureDiagram/.style={
		for tree={
			text depth = 0,
			parent anchor = south,
			child anchor = north,
			draw = drawColor,
			edge = {draw=drawColor},
			inner sep=3pt,
		}
	},
	/tikz/abstract/.style={
		fill = blue!85!cyan!5,
		draw = drawColor,
    font=\scriptsize
	},
	/tikz/concrete/.style={
		fill = blue!85!cyan!20,
		draw = drawColor,
    font=\scriptsize
	},
	mandatory/.style={
		edge label={node [mandatory] {} }
	},
	optional/.style={
		edge label={node [optional] {} }
	},
	or/.style={
		tikz+={
			\path (.parent) coordinate (A) -- (!u.children) coordinate (B) -- (!ul.parent) coordinate (C) pic[fill=drawColor, angle radius=\angleSize]{angle};
		}	
	},
	/tikz/or/.style={
	},
	alternative/.style={
		tikz+={
			\path (.parent) coordinate (A) -- (!u.children) coordinate (B) -- (!ul.parent) coordinate (C) pic[draw=drawColor, angle radius=\angleSize]{angle};
		}	
	},
	/tikz/alternative/.style={
	},
	/tikz/placeholder/.style={
	},
	collapsed/.style={
		rounded corners,
		no edge,
		for tree={
			fill opacity=0,
			draw opacity=0,
			l = 0em,
		}
	},
	/tikz/hiddenNodes/.style={
		midway,
		rounded corners,
		draw=drawColor,
		fill=white,
		minimum size = 1.2em,
		minimum width = 0.8em,
		scale=0.9
	},
	attributes/.style={
	    label={[align=left,font=\tiny]below:{\strut#1}}
	},
	attributesSide/.style={
	    label={[align=left,font=\tiny]right:{\strut#1}}
	},
	featureCard/.style={
	    label={[align=left,font=\scriptsize]below:{\strut#1}}
	},
	featureCardSide/.style={
	    label={[align=left,font=\scriptsize]right:{\strut#1}}
	},
	featureCardSideLeft/.style={
	    label={[align=left,font=\scriptsize,xshift=1mm,yshift=1mm]left:{\strut#1}}
	},
	dots/.style={
		label={[align=left,font=\scriptsize]left:{\strut#1}}
	}
}

\begin{document}

\title{Pseudo-Boolean d-DNNF Compilation for Expressive Feature Modeling Constructs}

\author{\IEEEauthorblockN{Chico Sundermann\IEEEauthorrefmark{1}\IEEEauthorrefmark{2},
    Stefan Vill\IEEEauthorrefmark{1}, Elias Kuiter\IEEEauthorrefmark{3}, Sebastian Krieter\IEEEauthorrefmark{2}, Thomas Thüm\IEEEauthorrefmark{2}, and Matthias Tichy\IEEEauthorrefmark{1}}
  \\
  \IEEEauthorblockA{\IEEEauthorrefmark{1} Ulm University, Germany}
  \IEEEauthorblockA{\IEEEauthorrefmark{2} TU Braunschweig, Germany}
  \IEEEauthorblockA{\IEEEauthorrefmark{3} University of Magdeburg, Germany}
}

\markboth{Technical Report}%
{Sundermann \MakeLowercase{\textit{et al.}}: Tackling Expressive Feature-Modeling Constructs with Pseudo-Boolean d-DNNF Compilation}


\maketitle

\begin{abstract}
  Configurable systems typically consist of reusable assets that have dependencies between each other.
  To specify such dependencies, feature models are commonly used.
  As feature models in practice are often complex, automated reasoning is typically employed to analyze the dependencies.
  Here, the de facto standard is translating the feature model to conjunctive normal form (CNF) to enable employing off-the-shelf tools, such as SAT or \ssat{} solvers.
  However, modern feature-modeling dialects often contain constructs, such as cardinality constraints, that are ill-suited for conversion to CNF.
  This mismatch between the input of reasoning engines and the available feature-modeling dialects limits the applicability of the more expressive constructs.
  In this work, we shorten this gap between expressive constructs and scalable automated reasoning.
  Our contribution is twofold:
  First, we provide a pseudo-Boolean encoding for feature models, which facilitates smaller representations of commonly employed constructs compared to Boolean encoding.
  Second, we propose a novel method to compile pseudo-Boolean formulas to Boolean d-DNNF.
  With the compiled d-DNNFs, we can resort to a plethora of efficient analyses already used in feature modeling.
  Our empirical evaluation shows that our proposal substantially outperforms the state-of-the-art based on CNF inputs for expressive constructs.
  For every considered dataset representing different feature models and feature-modeling constructs, the feature models can be significantly faster translated to pseudo-Boolean than to CNF.
  Overall, deriving d-DNNFs from a feature model with the targeted expressive constraints can be substantially accelerated using our pseudo-Boolean approach.
  For instance, the Boolean approach only scales for group cardinalities with up-to 15 features while pseudo-Boolean d-DNNF compilation can compile cardinalities with thousands of features.
  Furthermore, our approach is competitive on feature models with only basic constructs.
\end{abstract}


\section{Introduction}

Most aspects of modern life, such as transportation~\cite{EPP+:VaMoS23,KZK:LoCoCo10,OPS+:REJ17}, communication~\cite{BDC+:SETSS89,KCM:A-Mobile18}, and computers~\cite{LSB+:SPLC10,RNGB:SPLC24}, rely on configurable systems.
Such systems typically consist of multiple features that can be combined (\ie, configured) according to a set of constraints.
Variability languages are used to specify such constraints between features~\cite{BSL+:TSE13,BSE:MODEVAR19}.
For instance, a constraint may specify that a feature depends on another feature or that at least one of a group of features needs to be selected.
A common specification in variability languages are feature models~\cite{BSE:MODEVAR19}, which define features and constraints (\eg, dependencies) between them~\cite{B05}.

Industrial systems often come with thousands of constraints~\cite{KTM+:ESECFSE17,SBK+:SPLC24,LSB+:SPLC10,KZK:LoCoCo10}, which mandates the usage of automated analyses~\cite{OPS+:REJ17}.
For instance, common analyses are checking whether a given configuration is valid or computing the number of valid configurations~\cite{SKH+:AMAI24,BSRC10}.
The de facto standard for analyzing feature models is translating them to conjunctive normal form (CNF) then applying standardized reasoning engines, such as SAT~\cite{LGCR:SPLC15,B05,SKH+:AMAI24} or \ssat{} solvers~\cite{SNB+:VaMoS21,OGB+:TR19,KZK:LoCoCo10}.

There is a mismatch between constraint types in available variability languages and constraints supported by existing reasoning engines.
While research on feature-model analysis is mostly limited to Boolean logic~\cite{SKH+:SPLC24,LGCR:SPLC15}, the majority of variability languages support constraints that are hard to represent in CNF~\cite{BSE:MODEVAR19, FSTR:SPLC22, PNX+:FOSD11}.
Commonly considered examples for such constructs are group cardinality (\ie, select between $n$ and $m$ features from a group), feature cardinality (\ie, select a single feature between $n$ and $m$ times), or constraints over numeric attributes (\eg, overall power usage of features is limited), or non-Boolean features~\cite{BSE:MODEVAR19,BSF+:JSS25,gears,RFB+:MODEVAR22}.
For those constructs, translating to Boolean logic may lead to an exponential increase in formula size~\cite{BTS:SEFM19}.
In a survey of ter Beek~\etal~\cite{BSE:MODEVAR19}, 11 out 13 variability languages include at least one of such constructs.
Consequentially, it is often infeasible to represent and reason about feature models containing those constraints with the standard Boolean approaches~\cite{BTS:SEFM19}.

Such expressive constructs are also part of most languages used in practice.
In a survey on variability-modeling practices in the industry of Berger~\etal~\cite{BRN+:VaMoS13}, pure variants~\cite{pure-variants} was the most used off-the-shelf tool.
Their specification relies on various expressive constructs, including group cardinalities and feature attributes~\cite{RFB+:MODEVAR22}.
The tools following in popularity, namely Gears, FeatureIDE~\cite{MTS+17,SHE+:MODEVAR21} and DOPLER~\cite{DGR:AUSE11}, also support more expressive constructs.

Several studies also report the mandate of these constructs in real-world projects~\cite{HBH:SLE11,PNX+:FOSD11, BNR+:MODELS14, WAYL:QSIC13, BTS:SEFM19}.
Hubaux~\etal~\cite{HBH:SLE11} conducted a survey with four\footnote{The four companies come from the computer hardware, meeting management, and document management application domain~\cite{HBH:SLE11}.} companies and report that feature cardinalities and attribute constraints are desired.
Furthermore, expressive constructs, including group cardinalities, attribute constraints, and feature cardinalities, were also observed in other real-world case studies, such as the embedded system eCos~\cite{PNX+:FOSD11}, configurable jewelry-rings~\cite{BNR+:MODELS14}, video conference systems by Cisco~\cite{WAYL:QSIC13}, and university courses~\cite{BTS:SEFM19}.
\helppls{More case studies?}

Scalable solutions for various analyses on expressive constructs are lacking.
While solutions for checking satisfiability are also available for more expressive formats (\eg, SMT~\cite{MB:TACAS08,SSK+:VaMoS20} or CP~\cite{JRL:OSSICP08,BTR:SEKE05}), such techniques have not been yet extended for counting or enumeration.
Both of these operations are relevant for a plethora of feature-model analyses~\cite{SNB+:VaMoS21,GAT+:SPLC16,SKH+:AMAI24}.
Another issue hindering scalability is that many feature-model analyses rely on more than a single invocation of those complex computations~\cite{SKH+:AMAI24}.
Knowledge compilation is a strategy for automated reasoning which has recently attracted attention for feature-model analysis~\cite{SRH+:TOSEM24,SGRM:LPAIR18,HPF+:SOMET16}.
With knowledge compilation, the feature model is translated to a format (\eg, d-DNNF~\cite{D:AAAI02} or BDD~\cite{B18}) with beneficial properties~\cite{DM:JAIR02}.
These formats can be expensive to compile initially, but after this one-time initial effort, they support more efficient analysis techniques~\cite{DM:JAIR02}.
However, compilation for formats popular in feature-model analysis is currently strictly limited to Boolean formulas~\cite{MMBH:AAI12,D:AAAI02,LM:IJCAI17,SKH+:AMAI24,DHK:SPLC24}, which excludes the more expressive feature-modeling constructs.

In this work, we aim to shorten the gap between expressive feature-modeling constructs and suitable reasoning engines.
Our strategy to shorten this gap is twofold.
First, we provide pseudo-Boolean encodings for common feature-modeling constructs which are often more natural and substantially smaller than respective Boolean encodings.
In particular, we target constructs from the Universal Variability Language (UVL)~\cite{BSF+:JSS25}, namely all basic constructs and group cardinalities, feature cardinalities, and attribute constraints.
We use UVL as reference as it is widely employed~\cite{LSS+:SPLC23,BSF+:JSS25,RGS+:JSS24,FSRR:VaMoS21,MPFB:JSS23,SHE+:MODEVAR21,FFB+:MODEVAR22} and covers many common constructs~\cite{BSF+:JSS25,BSE:MODEVAR19}.
The advantage of pseudo-Boolean formulas is that they allow numeric constants, while still only having to manage Boolean variables.
Using this flexibility compared to CNFs facilitates encoding expressive constructs while profiting from the performance advantages of strictly Boolean variables.
This allows us to reuse established strategies from Boolean~\cite{T:SAT06,LM:IJCAI17} and pseudo-Boolean reasoning~\cite{CK:CADICS05,YM:AAAI24}.
Second, we introduce pseudo-Boolean d-DNNF compilation, which enables compiling pseudo-Boolean constraints to a Boolean d-DNNF.
The resulting d-DNNF can then be used to efficiently perform a plethora of relevant feature-model analyses~\cite{SRH+:TOSEM24,SKH+:AMAI24,SGRM:LPAIR18,BDD+:TR23}.
For instance, counting and enumeration configurations can be performed in linear time (\wrt d-DNNF size) on a d-DNNF~\cite{DM:JAIR02,SKH+:AMAI24}.
Overall, we provide the following contributions in this work:
\begin{itemize}
  \item We present \emph{pseudo-Boolean encodings} for several common feature-modeling constructs, including the expressive constructs group cardinality, feature cardinality and attribute constraints (\autoref{sec:pseudoencoding}).
  \item We propose a core algorithm and optimizations for \emph{compiling pseudo-Boolean constraints to d-DNNF} (\autoref{sec:compilation}).
  \item We provide \emph{publicly available tooling} for both pseudo-Boolean encoding and pseudo-Boolean d-DNNF compilation.
  \item We examine the advantages of our approach regarding both encoding and compilation over the state-of-the-art with a \emph{large-scale empirical evaluation}  (\autoref{sec:evaluation}).
\end{itemize}



\section{Background} \label{sec:background}

In this section, we introduce the background required for understanding the main concepts in this work, namely feature models, automated reasoning, d-DNNF compilation, and pseudo-Boolean logic.

\subsection{Feature Models}
\emph{Feature models} are a commonly used formalism to specify the set of valid configurations for a configurable system~\cite{KCHNP90}.
Typically, a feature model consists of a feature hierarchy, denoting parent-child relationships with additional cross-tree constraints~\cite{KCHNP90,BSRC10}.
Often, non-functional attributes can be attached to features (\eg, a price)~\cite{SSA:ESECFSE17,CSHL:ICSE13,SSK+:VaMoS20,BSF+:JSS25,BSE:MODEVAR19}.
However, the types of parent-child relationships and constraints that are supported vastly differ in literature~\cite{BSE:MODEVAR19}.

\autoref{table:fmtaxonomy} shows a taxonomy of different feature-modeling classes.
The taxonomy is inspired by the language levels used in the Universal Variability Language (UVL)~\cite{BSF+:JSS25}.
We use UVL as reference since UVL is widely used~\cite{SHE+:MODEVAR21,LSS+:SPLC23,RGS+:JSS24,MPFB:JSS23,FSRR:VaMoS21} and covers many common constructs~\cite{BSF+:JSS25,BSE:MODEVAR19}.
Furthermore, UVL language levels and our taxonomy aim to categorize the feature-modelling based on complexity for reasoning engines.
For instance, basic feature models can be translated to CNF within poly-time and poly-space~\cite{KKS+:ASE22}, which is the de facto standard input for SAT~\cite{B:JSAT08,LP:JSAT10} or \ssat{}~\cite{T:SAT06,SRSM:IJCAI19,BSB:SAT15} solvers.
In contrast, feature models with different feature types require more powerful reasoning (\eg, SMT~\cite{MB:CACM11,SSK+:VaMoS20,LSS+:SPLC23}).
Overall, we differentiate between four different classes, namely basic $\mathbb{B}$, cardinality $\mathbb{C}$, attribute constraints $\mathbb{A}$, and feature types $\mathbb{T}$.
The classes differ in the supported group types (\ie, hierarchical parent-child relationships), types of cross-tree constraints (\eg, propositional), and feature types (\eg, Boolean or numeric).
In this work, we consider all constructs from $\mathbb{C}$, $\mathbb{A}$, and $\mathbb{T}$ as \emph{expressive} constructs.
Furthermore, we use $G_{\mathbb{X}}$ to denote the group types introduced by the class $\mathbb{X}$ and $C_{\mathbb{X}}$ types of cross-tree constraints introduced by the class $\mathbb{X}$.
For instance, the cardinality class $\mathbb{C}$ introduces the group types $G_{\mathbb{C}} = \{\text{group cardinality, feature cardinality}\}$.

\begin{table*}
  \centering
  \caption{Feature-Modeling Construct Taxonomy Inspired By UVL Language Levels~\cite{BSF+:JSS25}}
  \label{table:fmtaxonomy}
  \begin{tabular}{llll}
    \toprule
    Class                              & Introduced Groups ($G_{\mathbb{X}}$)              & Introduced Cross-tree Constraints ($C_{\mathbb{X}}$)              & Introduced Feature Types \\
    \midrule
    Basic $\mathbb{B}$                 & \{Alternative, or, mandatory, optional\}          & Propositional ($\land, \lor, \neg, \Leftrightarrow, \Rightarrow)$ & \{Boolean\}              \\
    Cardinality $\mathbb{C}$           & \{\text{group cardinality, feature cardinality}\} & -                                                                                            \\
    Attribute Constraints $\mathbb{A}$ & -                                                 & attribute terms                                                   & -                        \\
    Feature Types $\mathbb{T}$         & -                                                 & \text{feature terms}                                              & \{numeric, string\}      \\
    \bottomrule
  \end{tabular}
\end{table*}

\begin{figure}[htbp]
  \centering
  \begin{forest}
    featureDiagram
    [Robot Vacuum,concrete
    [Maps,concrete,optional
    ]
    [Obstacle Detection,concrete, mandatory
      [Sensor, concrete, or
      ]
      [
        AI, concrete, or
      ]
      [Physical, concrete
      ]
    ]
    [Mop Mode, concrete, optional
    ]
    [Camera,concrete,optional
    ]
    [Extra Storage,concrete,mandatory
    [
    Dust storage, concrete, alternative
    ]
    [
    Water storage, concrete
    ]
    ]
    ]
    \matrix [draw=drawColor,anchor=north west, font=\footnotesize] at (-3,-2) {
      \node [label=center:\underline{Legend:}] {}; \\
      \node [concrete,label=right:Feature] {};        \\
      \node [mandatory,label=right:Mandatory] {};     \\
      \node [optional,label=right:Optional] {};       \\
      \filldraw[drawColor] (0.1,0) - +(-0,-0.2) - +(0.2,-0.2)- +(0.1,0);
      \draw[drawColor] (0.1,0) -- +(-0.2, -0.4);
      \draw[drawColor] (0.1,0) -- +(0.2,-0.4);
      \fill[drawColor] (0,-0.2) arc (240:300:0.2);
      \node [or,label=right:Or Group] {};             \\
      \draw[drawColor] (0.1,0) -- +(-0.2, -0.4);
      \draw[drawColor] (0.1,0) -- +(0.2,-0.4);
      \draw[drawColor] (0,-0.2) arc (240:300:0.2);
      \node [alternative,label=right:Alternative Group] {}; \\};
    \matrix[font=\footnotesize] [anchor=north, align=center] at (3, -2) {
      \node {(AI $\lor$ Maps) $\Rightarrow$ Camera}; \\
      \node {Mop Mode $\Rightarrow$ Water Storage};  \\
    };
  \end{forest}
  \caption{Basic Feature Model}
  \label{figure:background:basicexample}
\end{figure}
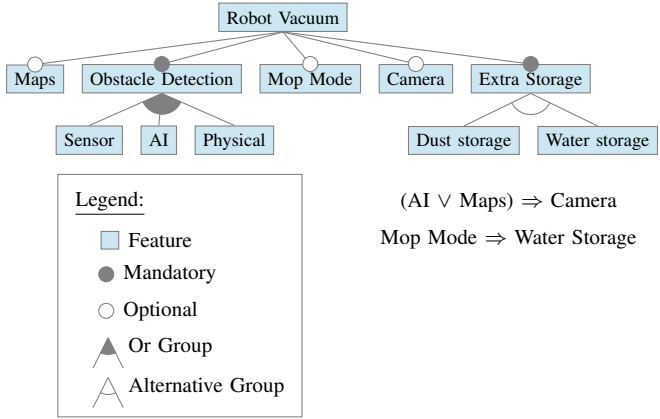

\emph{Basic} feature models are commonly considered, especially in the context of automated reasoning~\cite{MTS+17,GHF+:SPLC23,ACLF:SCP13,LGCR:SPLC15,SKH+:AMAI24}.
\autoref{figure:background:basicexample} showcases a basic feature model specifying the valid configurations of a robot vacuum product line.
Each \featurestyle{robot vacuum} requires some kind of \featurestyle{obstacle detection} and \featurestyle{extra storage} (\emph{mandatory} flag).
Furthermore, the robots may have \featurestyle{maps}, a \featurestyle{mop mode}, and a \featurestyle{camera} (\emph{optional} flag).
For the obstacle detection, users may choose between one or more from \featurestyle{sensor}, \featurestyle{AI}, and \featurestyle{physical} (\emph{or} relation).
The extra storage can be either a \featurestyle{dust storage} or a \featurestyle{water storage} (\emph{alternative} relation).
In addition to the hierarchy, two cross-tree constraints denote that \featurestyle{AI} and \featurestyle{maps} rely on having a \featurestyle{camera} and that \featurestyle{mop mode} requires a \featurestyle{water storage}.
In basic feature models, the typical parent-child relationships are \emph{optional}, \emph{mandatory}, \emph{alternative}, and \emph{or}~\cite{MTS+17,GHF+:SPLC23,BSF+:JSS25}.
Furthermore, cross-tree constraints are limited to propositional logic with common operators.

\begin{figure}[htbp]
  \centering
  \begin{forest}
    featureDiagram
    [Robot Vacuum,concrete
    [Maps,concrete,optional,
    attributes={\{cost = 7\}}
    ]
    [Obstacle Detection,concrete, mandatory
      [Sensor, concrete, edge label={node[midway,auto,xshift=10.5mm,yshift=2.5mm]{\scriptsize [1..2]}},
        attributes={\{cost = 2\}}
      ]
      [
        AI, concrete,
        attributes={\{cost = 4\}}
      ]
      [Physical, concrete,
        attributes={\{cost = 3\}}
      ]
    ]
    [Mop Mode, concrete, optional,
    attributes={\{cost = 2\}}
    ]
    [Camera,concrete,optional,
    attributes={\{cost = 2\}}
    ]
    [Extra Storage,concrete,mandatory, featureCardSide={[1..3]}
      [
        Dust storage, concrete, alternative,
        attributes={\{cost = 3, \\ space = 2\}}
      ]
      [
        Water storage, concrete,
        attributes={\{cost = 5, \\ space = 4\}}
      ]
    ]
    ]
    \matrix [anchor=north] at (current bounding box.south) {
      \node [placeholder] {}; \\
    };
    \matrix[font=\footnotesize] [anchor=north] at (current bounding box.south) {
      \node {(AI $\lor$ Maps) $\Rightarrow$ Camera};                 \\[-2pt]
      \node {Mop Mode $\Rightarrow$ Water Storage};                  \\[-2pt]
      \node {\textbf{sum}(cost) $<$ 15};                         \\[-2pt]
      \node {"Dust storage".space + "Water storage".space $\leq$ 6}; \\
    };
  \end{forest}
  \caption{Feature Model Including Expressive Constructs}
  \label{figure:background:expressiveexample}
\end{figure}
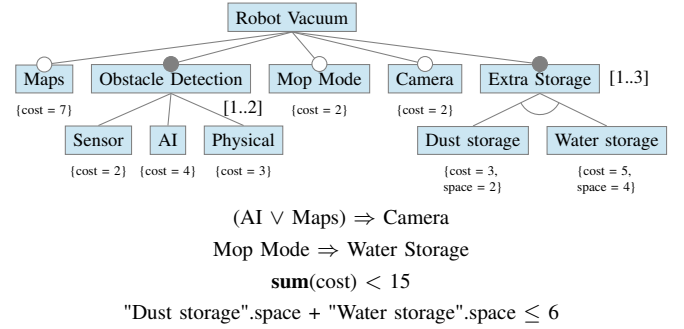

However, the majority of available variability languages also support at least some more expressive constructs, such as cardinalities, attributes and constraints over them, or typed features~\cite{BSE:MODEVAR19,gears,RFB+:MODEVAR22}.
\autoref{figure:background:expressiveexample} shows an extension of the feature model shown in \autoref{figure:background:basicexample} with constructs from $\mathbb{B}$, $\mathbb{C}$, and $\mathbb{A}$.
For obstacle detection, due to a limited number of connections, only one or two features can be selected (\emph{group cardinality} [1..2]).
Between one and three extra storage components can now be configured (feature cardinality [1..3]).
In addition, the features now have non-functional properties, called \emph{feature attributes}~\cite{SSA:ESECFSE17}.
Here, most features have a cost and the storage components also have an attribute indicating the space they require.
These attributes can be used for informational purposes or to specify constraints over them.
In our example, the overall cost is limited to 15 and the overall space for storages is limited to 6.
Various interpretations of feature attribute semantics have been discussed~\cite{KOD:SCP13,BSF+:JSS25,SSK+:VaMoS20,BBGA:SPLC15,OGRT:SLE15}.
We follow UVL~\cite{BSF+:JSS25} regarding the semantics of feature attributes.
There is one important difference between feature attributes and typed features: attributes cannot be configured individually (\ie, their value is constant when deriving configurations)~\cite{BSF+:JSS25,SSA:ESECFSE17} \helppls{More cite}.
Hence, features are variables while attributes are constants which are typically easier to handle for reasoning engines.
Formally, we consider attributes $a_i = (f_i, v_i)$ as always attached to a feature $f_i$.
In a configuration, their value is $v_i$ if its feature $f_i$ is selected and zero otherwise.
We discuss the semantics of the presented relationships and constraints in \autoref{sec:pseudoencoding}.
Nevertheless, we refer to related work on the structure and semantics of feature models for more details~\cite{KCHNP90,B05,BSF+:JSS25}.
\todo{Maybe discuss different types of attributes}

\subsection{Logic-Based Reasoning} \label{subsec:background:logic}
The de facto standard for feature-model analysis is translating the feature model to Boolean logic~\cite{SKH+:AMAI24,B05,BSRC10,MKR+:ICSE16,PAP+:ICST19,LGCR:SPLC15} and then invoke off-the-shelf solvers~\cite{MWC:SPLC09,T:SAT06,MTS+17,LP:JSAT10,GHF+:SPLC23}.
Here, the features are mapped to Boolean variables and the parent-child relationships and cross-tree constraints translated to Boolean formulas, typically in conjunctive normal form (CNF)~\cite{HR04}.
A CNF $C$ is a single conjunction of clauses $C = (C_1 \land \ldots \land C_n)$.
Each clause $C_i$ is a disjunction of literals $C_i = (l_{i,1} \lor \ldots \lor l_{i,m})$.
For basic feature models $\mathbb{B}$, translations to CNF are well known and commonly employed~\cite{MTS+17,CW:SPLC07,KKS+:ASE22}.
However, constructs from the expressive classes $\mathbb{C}$, $\mathbb{A}$, and $\mathbb{T}$ are not straightforward to translate to CNF.
In general, there is a lack of scalable automated reasoning techniques for those expressive constructs.
\todo{Maybe discuss conversions for cardinality constraints already here}

\subsection{d-DNNF Compilation} \label{subsec:background:ddnnfcompile}
\emph{Knowledge compilation} refers to compiling an original formula (\eg, representing a feature model) to another format with beneficial properties regarding runtime of complex operations~\cite{DM:JAIR02}.
There are various knowledge-compilation target languages that each offer varying tradeoffs between (1) the complexity to compile and (2) operations that are possible in polynomial time on the compiled artifact~\cite{DM:JAIR02}.
Typical operations are checking satisfiability, model counting, enumerating solutions, or algebraic operations~\cite{SKH+:AMAI24,DM:JAIR02}.

The \emph{deterministic decomposable negation normal form} (d-DNNF) is a knowledge compilation target language that has gained traction in feature-model analysis over the past years~\cite{SRH+:TOSEM24,BDD+:TR23,SGRM:LPAIR18,SLT:ASE24}. d-DNNFs support model counting, checking satisfiability, and enumerating solutions within linear time with respect to the d-DNNF's size~\cite{DM:JAIR02}, which can be used to accelerate a plethora of feature-model analyses~\cite{SKH+:AMAI24}.
None of these operations can be computed in polynomial time on a CNF~\cite{DM:JAIR02}.
The linear runtime operations for d-DNNFs are enabled by the properties \emph{determinism} and \emph{decomposability}. A formula is deterministic if all disjuncts $d_i$ in a disjunction $D = d_1 \vee \ldots d_n$ are pairwise logically contradicting (\ie, $\forall i,j : i \neq j: d_i \land d_j \equiv \bot$). A formula is decomposable if no conjunct $c_i$ in a conjunction $C = c_1 \ldots c_n$ shares variables with another conjunct (\ie, $\forall i,j : i \neq j: vars(c_i) \cup vars(c_j) = \emptyset$).
\todo{TT: maybe add example on how to count}

Modern d-DNNF compilation is based on the exhaustive Davis-Putnam-Logemann-Loveland (DPLL) algorithm~\cite{LM:IJCAI17,MMBH:AAI12,D:AAAI02}.
Here, the solution space of the formula is explored by recursively assigning variables until every solution is found. \autoref{algorithm:booleanddnnf} shows the base procedure to compile a given CNF to d-DNNF. There are two main aspects to consider. First, while the formula is not yet satisfied nor unsatisfied, variables are assigned recursively. The subformulas $F|_v$ and $F|_{\neg v}$ resulting from setting $v$ to true and false are recursively compiled to d-DNNF and combined to a disjunction $(F|_v \wedge v) \vee (F|_{\neg v} \wedge \neg v)$. Note that this disjunction is equivalent to $F$ and per construction deterministic as every leftside solution contains $v$ and every rightside solution contains $\neg v$. Second, if the formula can be split into subformulas that share no variables (\ie, components), those components are then processed separately. The compiled sub d-DNNFs can then be conjuncted. Note that the conjunction is equivalent to $F$ and decomposable by construction. Following this procedure, we acquire a semantically equivalent d-DNNF since every conjunction is decomposable and every disjunction is deterministic.

\begin{algorithm}
  \caption{Boolean d-DNNF Compilation}
  \label{algorithm:booleanddnnf}
  \begin{algorithmic}[1]
    \STATE \textbf{Input:} CNF $F$
    \STATE \textbf{Output:} d-DNNF $D$
    \STATE \textbf{Procedure:} compile($F$)
    \IF {satisfied($F$)}
    \STATE \textbf{return} $\boxed{\top}$
    \ELSIF {!satisfied($F$)}
    \STATE \textbf{return} $\boxed{\bot}$
    \ENDIF
    \STATE components $\gets$ identifyDisconnectedComponents($F$)
    \STATE subs $\gets$ [ ]
    \FOR {$C$ : components}
    \IF {sat($C$)}
    \STATE \textbf{continue}
    \ENDIF
    \STATE subs.push(compile($F|_v \wedge v) \vee \text{compile}(F|_{\neg v} \wedge \neg v$))
      \ENDFOR
      \STATE \textbf{return} $\bigwedge\limits_{S \in \text{subs}} S$
  \end{algorithmic}
\end{algorithm}

\subsection{Pseudo-Boolean Logic} \label{subsec:background:pseudoboolean}
A linear pseudo-Boolean constraint is an inequality following the structure shown in \autoref{equation:pseudoboolean}~\cite{CK:CADICS05,HR:OR69}.
Here, $k_i$ and $b$ are constants while $x_i$ are Boolean variables with $x_i \rightarrow \{0,1\}$, where $0$ corresponds to false and $1$ to true.
The set $\text{vars}(F)$ denotes the variables considered in the formula.
The constants $k_i$ are called \emph{factors} and the rightside $b$ is called \emph{degree}~\cite{CK:CADICS05}.
Each pseudo-Boolean formula is a single conjunction over such constraints.
Consequentially, pseudo-Boolean formulas are a generalization of CNFs.
A CNF clause $(x_1 \vee \ldots \vee x_n$) can be represented as $\sum_{i=1}^n 1 \cdot x_i \geq 1$.

\begin{equation}
  \sum_{i=1}^n k_i \cdot x_i \geq b \text{ with } k_i, b \in \mathbb{Z}, x_i \in \text{vars}(F)
  \label{equation:pseudoboolean}
\end{equation}

The advantage of pseudo-Boolean is the added expressiveness over CNFs with numerical constants while preserving strictly Boolean variables~\cite{CK:CADICS05,HR:OR69}.
With the added expressiveness, certain constructs can be encoded more sparsely (\ie, with fewer literals).
In \autoref{sec:pseudoencoding}, we showcase these advantages by encoding different feature-modeling constructs in pseudo-Boolean logic.
By having strictly Boolean variables, we can reuse strategies (\cf \autoref{sec:compilation}) that have been optimized over decades of research on Boolean~\cite{B:JSAT08,T:SAT06,LM:IJCAI17} and pseudo-Boolean~\cite{CK:CADICS05,YM:AAAI24,HK:JSAT06} reasoning.
\todo{TT: Maybe extend,  ek: possibilities for extensions: what are general application areas of pseudo-Boolean constraints?, mt: relation zu ILP and BIP}

\section{Pseudo-Boolean Feature Model Encoding} \label{sec:pseudoencoding}

\todo{Rename variables}

To use the benefits of pseudo-Boolean logic for feature modeling, we first need encodings of commonly occuring constructs.
While Boolean encodings for feature models are well researched, we only found a Bachelor's thesis by Hennerberg~\cite{Henneberg11} addressing encodings as pseudo-Boolean formulas, but they are limited to basic constructs, group cardinalities, and sums over attributes.
Furthermore, their group cardinality encoding produces faulty results if the parent is deselected~\cite{Henneberg11}.
Hence, we provide an extended collection of pseudo-Boolean encodings.
Note that there are typically multiple valid encodings for a feature-modeling construct.
In this work, for each Boolean and pseudo-Boolean, we limit ourselves to one \textit{semantically equivalent} encoding per construct.

Various feature-modeling constructs have been considered with varying degrees of adoption in practice~\cite{BSE:MODEVAR19}.
With our approach, we limit ourselves to constructs from the Universal Variability Language (UVL)~\cite{BSF+:JSS25} for now.
We use UVL as reference since UVL is widely used~\cite{SHE+:MODEVAR21,LSS+:SPLC23,RGS+:JSS24,MPFB:JSS23,FSRR:VaMoS21} and covers many common constructs~\cite{BSF+:JSS25}.
Furthermore, the design decisions including supported constructs are result of a community survey with researchers and practitioners~\cite{SFE+:SPLC21} which increases the chances for the constructs to be relevant.
With pseudo-Boolean encoding, we aim to find a sweet spot that provides more efficient encodings for several constructs compared to CNF (\cf \autoref{subsec:encoding:basic} and \autoref{subsec:encoding:expressive}), but can still capitalize on the performance advantages of purely Boolean variables.
Hence, we target the first three classes of our taxonomy, namely basic $\mathbb{B}$, cardinality $\mathbb{C}$, and attribute constraints $\mathbb{A}$.
The taxonomy classes $\mathbb{B}$, $\mathbb{C}$, and $\mathbb{A}$ cover all UVL language levels but \emph{typed} features~\cite{BSF+:JSS25}.

\todo{Maybe integrate robot vacuum example here more often}


\subsection{Encoding Basic Feature Models} \label{subsec:encoding:basic}
The goal of employing pseudo-Boolean logic for feature models is mainly to provide more efficient encodings (\wrt size and translation runtime) for additional expressive constructs.
Nevertheless, to translate feature models in practice we initially require encodings for basic constructs.
\autoref{table:encoding:basichierarchy} shows Boolean and pseudo-Boolean encodings for the hierarchy group types in features models from the basic class $\mathbb{B}$. Here, $\mathit{grp}^p_{c_1..c_n}$ describes the relationship $\mathit{grp} \in \{\mathit{opt}, \mathit{mand}, \mathit{or}, \mathit{alt}\}$ between parent $p$ and its children $\{c_1,\ldots, c_n\}$.
Note that a feature $c_i$ can only be child feature in one group, but a parent feature $p$ may have multiple groups.
For the Boolean encoding, we use FeatureIDE's\footnote{\url{https://github.com/FeatureIDE/FeatureIDE}. FeatureIDE is a widely used framework for software-product line development~\cite{MTS+17}.}~\cite{MTS+17} translations as reference, as they are commonly used~\cite{BDD+:TR23,KAT:GPCE16,KKS+:ASE22,FSRR:VaMoS21} and in line with other typical definitions from the literature. \helppls{Any more pointers that are not from us?}
For the pseudo-Boolean encoding, our translations are similar to the ones provided by Henneberg~\cite{Henneberg11}.

\begin{table}[htbp]
  \centering
  \caption{Encodings for Basic Feature Model Hierarchy}
  \label{table:encoding:basichierarchy}
  \footnotesize
  \addtolength{\tabcolsep}{-0.6em}
  \begin{tabular}{l l r}
    \toprule
    \textbf{Constraint}            & \textbf{Boolean}                                                                          & \textbf{Pseudo-Boolean}                \\
    \midrule
    $\mathit{opt}^{p}_{c_1..c_n}$  & $\bigwedge^n_{i=1} (c_i \Rightarrow p)$                                                       & $n \cdot p + \sum^n_{i=1} -c_i \geq 0$ \\[3pt]
    $\mathit{mand}^{p}_{c_1..c_n}$ & $\bigwedge^{n}_{i=1} (p \Leftrightarrow c_i)$                                               & $n \cdot p + \sum^n_{i=1} -c_i = 0$    \\[3pt]
    $\mathit{or}^{p}_{c_1..c_n}$   & $p \Leftrightarrow (\bigvee^{n}_{i=1} c_i)$                                                 & $n \cdot p + \sum^n_{i=1} -c_i \geq 0$ \\[1pt]
                                   &                                                                                           & $ -p + \sum^n_{i=1} c_i \geq 0$        \\[3pt]
    $\mathit{alt}^{p}_{c_1..c_n}$  & $ p \Leftrightarrow (\bigvee^n_{i=1} c_i \land \bigwedge_{i<j} (\neg c_i \vee \neg c_j))$ & $ p + \sum^n_{i=1} -c_i = 0$           \\
    \bottomrule
  \end{tabular}
\end{table}

\paragraph{Optional Features}
For \emph{optional} features, no constraints are imposed on the child features when the parent is selected. Hence, an optional group only enforces that the selection of a child mandates the selection of the parent feature. \autoref{table:encoding:basichierarchy} shows the Boolean and pseudo-Boolean encoding for an optional group $opt^p_{c_1..c_n}$. The pseudo-Boolean constraint is only violated (\ie, left side $\geq$ 0) if a child is selected but the parent is not.
Note that $n \cdot p$ is a valid summand in a pseudo-Boolean constraint as $n$ is a constant and $p$ is a variable.

\paragraph{Mandatory Features}
\emph{Mandatory} features need to be selected if the parent is selected. The pseudo-Boolean constraint for $\mathit{mand}^{p}_{c_1..c_n}$ given in \autoref{table:encoding:basichierarchy} is violated if the parent but not all its children are selected.
Note that the provided pseudo-Boolean constraint does not exactly match the definition from \autoref{subsec:background:pseudoboolean} (\ie, $=$ instead of $\geq$), but can be straightforwardly normalized to the desired form.
We discuss normalizations we perform at the end of this section.

\paragraph{Or Group}
If the parent of an \emph{or} group $\mathit{or}^p_{c_1,..c_n}$, is selected, at least one of its child features need to be selected.
To specify $\mathit{or}^p_{c_1,..c_n}$, we use two pseudo-Boolean constraints that ensure (1) no $c_i$ is selected while $p$ is not and (2) selecting $p$ requires selecting at least one $c_i$.

\paragraph{Alternative Group}
If the parent of an \emph{alternative group} $\mathit{alt}^{p}_{c_1..c_n}$ is selected, exactly one of its child features needs to be selected. The pseudo-Boolean constraint for $\mathit{alt}^{p}_{c_1..c_n}$ is only satisfied if the parent $p$ and exactly one child $c_i$ (\ie, $1-1 =0$) is selected or no parent and no child (\ie, $0-0 =0$) is selected.

\paragraph{Boolean Cross-Tree Constraints}
To translate Boolean cross-tree constraints to pseudo-Boolean logic, we can use well-known translations to CNF and then translate the resulting clauses to pseudo-Boolean constraints.
In particular, we convert a clause $(x_1 \vee \ldots \vee x_n)$ to the pseudo-Boolean constraint $\sum_{i=1}^{n} 1 \cdot x_i \geq 1$.
The benefit is that we employ available techniques and tools for CNF translation~\cite{KKS+:ASE22}.

By combining these translations, we can represent basic feature models in pseudo-Boolean logic.
For basic feature models, most constructs can be represented as formulas with similar sizes in Boolean and pseudo-Boolean. The exception are alternatives $\mathit{alt}^{p}_{c_1..c_n}$ which induce considerable difference as we require $\mathcal{O}(n^2)$ clauses in a CNF, but only a single pseudo-Boolean constraint.

\subsection{Encoding Expressive Constraints} \label{subsec:encoding:expressive}
\autoref{table:encoding:expressive} shows Boolean and pseudo-Boolean encodings for constructs in the taxonomy classes cardinality $\mathbb{C}$ and attribute constraints $\mathbb{A}$.
As reference for the Boolean encoding, we use UVL conversion strategies~\cite{SVT+:SPLC23,BSF+:JSS25} which translate the expressive constructs to propositional constraints.
We employ the UVL conversion strategies as (1) they are part of the standard UVL tooling and (2) each conversion produces semantically equivalent formulas~\cite{BSF+:JSS25,SVT+:SPLC23}.

\begin{table}[htbp]
  \centering
  \caption{Encodings for Expressive Feature-Modeling Constructs}
  \label{table:encoding:expressive}
  \footnotesize
  \begin{tabular}{l l r}
    \toprule
    \textbf{Construct}                           & \textbf{Boolean}                                   & \textbf{Pseudo-Boolean}                \\
    \midrule
    $[a..b]^p_{c_1..c_n}$                        & enum($[a..b]^p_{c_1..c_n}$)                        & $n \cdot p + \sum^n_{i=1} -c_i \geq 0$ \\[1pt]
                                                 &                                                    & $-a \cdot p + \sum^n_{i=1} c_i \geq 0$ \\[1pt]
                                                 &                                                    & $\sum^n_{i=1} -c_i \geq -b$            \\[4pt]
    $f[a..b]$                                    & enum($[a..b]^{cr}_{c_1,..,c_b}$)                   & $pb([a..b]^{cr}_{c_1,..,c_b}$)         \\[3pt]
    $\phi(A) \, \termoppplaceholder \, \psi(A')$ & enum($\phi(A) \, \termoppplaceholder \, \psi(A')$) & \autoref{table:encoding:arithmetic}    \\
    \bottomrule
  \end{tabular}
\end{table}

\paragraph{Group Cardinality}
A group cardinality $[a..b]^p_{c_1..c_n}$ specifies that if parent $p$ is selected, between $a$ and $b$ of its children $c_i$ need to be selected.
For Boolean, we use the UVL conversion strategy which enumerates the valid assignments over the children $c_1..c_n$ that satisfy the group cardinality.
For instance, $[2..3]^p_{\{c_1, c_2, c_3\}}$ would be encoded as $(c_1 \land c_2 \land c_3) \lor (c_1 \land \bar{c_2} \land \bar{c_3})\lor (\bar{c_1} \land c_2 \land \bar{c_3}) \lor (\bar{c_1} \land \bar{c_2} \land c_3)$.
Note that several alternative encodings for at-least-k and at-most-k constraints have been suggested in the literature~\cite{BTS:SEFM19,S:CP05, BB:CP03,KK07}. We discuss those encodings and their suitability for d-DNNF compilation in \autoref{sec:relatedwork}.
The enumeration strategy is also employed by UVL for the remaining expressive constructs~\cite{BSF+:JSS25}.
With our pseudo-Boolean encoding shown in the first rows of \autoref{table:encoding:expressive}, we use three constraints that (1) ensure that child features $c_1,..,c_n$ can only be selected if parent $p$ is selected, (2) if $p$ is selected, at least $a$ child features are selected, and (3) at most $b$ child features are selected.
Henneberg~\cite{Henneberg11} also provides an encoding for group cardinalities, but his proposal may produce incorrect results if the parent is deselected.

\paragraph{Feature Cardinality}
A feature cardinality $f[a..b]$ specifies that a feature $f$ may be selected between $a$ and $b$ times~\cite{BSF+:JSS25,CK05}.
It is important to consider that for every incarnation $f_{i \in [1,b]}$ of $f$, the subtree spanned by ancestors of $f$ needs to be cloned as every incarnation $f_i$ can be configured individually~\cite{CK05,BSF+:JSS25,GWSL:VaMoS24}.
A feature cardinality can therefore be considered as group cardinality $[a..b]^{cr}_{f_1,..,f_b}$ over these clones $f_{i \in [1,b]}$ with a new artificial cardinality root feature $cr$.
This semantically equivalent interpretation is implicitly used by UVL~\cite{BSF+:JSS25} and explicitly used for our pseudo-Boolean encoding.
We use the previously presented strategy to convert the resulting group cardinality ($pb([a..b]^{cr}_{c_1,..,c_b}$) in \autoref{table:encoding:expressive}).
We manage cross-tree constraints containing cloned variables in the same way as UVL~\cite{BSF+:JSS25}.

\paragraph{Numeric Attribute Constraints}
In UVL, numeric attribute constraints are terms $\phi(A) \, \termoppplaceholder \, \psi(A')$ over constants and a set of feature attributes $A$ with $\termoppplaceholder \in \{=, \neq, \geq, \leq, >, <\}$.
A term $\phi(A)$ is an  arbitrary expression over attributes $a_i \in A$ and arithmetic operations (\ie, $+, -, \div, \cdot$).
An example is shown in \autoref{figure:background:expressiveexample} where the space of storage units is limited ("Dust storage".space + "Water storage".space $\leq$ 6).
In that example, we have $\phi(A) =$ ("Dust storage".space + "Water storage".space), $\psi(A') = 6$, and $\termoppplaceholder = \, \leq$.
Attributes $a_i = (f_i, v_i)$ resolve to their value $v_i$ if the corresponding feature $f_i$ is selected and to $0$ if $f_i$ is deselected.
UVL also contains two aggregate functions: sum() and avg().
However, since both can be expanded (\eg, sum($a$) to a sum over all instances $a_i$ of that attribute), we do not explicitly consider them here.
For the Boolean encoding, we also use the UVL conversion strategies, which enumerate satisfiable assignments over the features $f_i$ corresponding to the respective attributes $a_i \in A \cup A'$~\cite{BSF+:JSS25}.
For the pseudo-Boolean encoding, we propose an approach that \emph{recursively transforms subterms} to pseudo-Boolean logic depending on the connecting operator. We show the different transformations in \autoref{table:encoding:arithmetic}.

\begin{table}[htbp]
  \centering
  \caption{Pseudo-Boolean Expression Encodings}
  \label{table:encoding:arithmetic}
  \small
  \begin{tabular}{l l}
    \toprule
    Term                                                     & Target Encoding                                                                   \\
    \midrule
    $c_i$                                                      & $c_i$                                                                               \\[3pt]
    $a_i = (v_i, f_i)$                                       & $v_i \cdot f_i$                                                                   \\[3pt]
    $(c_1 + \sum^n_{i=1} k_i \cdot x_i) \, \bm{+}$           & $(c_1 + c_2) + \sum^n_{i=1} k_i \cdot x_i$                                        \\[1pt]
    $(c_2 + \sum^m_{i=1} l_i \cdot y_i)$                     & $+ \sum^m_{i=1} l_i \cdot y_i$                                                    \\[7pt]
    $(c_1 + \sum^n_{i=1} k_i \cdot x_i) \,  \bm{-} $         & $(c_1 - c_2) + \sum^n_{i=1} k_i \cdot x_i$                                        \\[1pt]
    $(c_2 + \sum^m_{i=1} l_i \cdot y_i)$                     & $+ \sum^m_{i=1} -l_i \cdot y_i$                                                   \\[7pt]
    $(c_1 + \sum^n_{i=1} k_i \cdot x_i) \, \, \, \bm{\cdot}$ & $(c_1 \cdot c_2) + \sum^n_{i=1} c_2 \cdot k_i \cdot x_i$                          \\[1pt]
    $ (c_2 + \sum^m_{i=1} l_i \cdot y_i) $                   & $ + \sum^n_{i=1} \sum^m_{j=1} k_i \cdot l_j \cdot z_{i,j} $                       \\[1pt]
                                                             & $+ \sum^m_{i=1} c_1 \cdot l_i \cdot y_i$                                          \\[7pt]
    $(c_1 + \sum^n_{i=1} k_i \cdot x_i) \, \, \div $         & $\sum_{s \in S} \frac{c_1}{c_2 + \sum_{i \in s} l_i} \cdot e_s$                   \\[1pt]
    $ (c_2 + \sum^m_{i=1} l_i \cdot y_i)$                    & $+ \sum^n_{j=1} \sum_{s \in S} \frac{k_j}{c_2 + \sum_{i \in s} \, l_i} \cdot e_s$ \\[3pt]
    \bottomrule
  \end{tabular}
\end{table}

For unary terms, \emph{constants} $c_i$ and \emph{feature attributes} $a_i$, the transformation is straightforward.
To translate the binary expressions, we always assume that both sides are already in pseudo-Boolean format (\ie, $(c_1 + \sum^n_{i=1} k_i \cdot x_i)$).\footnote{The constraint does not conflict with the structure shown in \autoref{equation:pseudoboolean} as the additional constant $c_1$ can bet set of against the degree.}
Otherwise, we can just recursively translate them using the same procedure.
For \emph{addition} and \emph{subtraction}, joining the sums and constants already produces a valid pseudo-Boolean expression.
When resolving the \emph{multiplication} of both sums, we produce $\sum^n_{i=1} \sum^m_{j=1} k_i \cdot l_j \cdot x_i \cdot y_j$, which is not in pseudo-Boolean format since $x_i$ and $y_j$ are both variables.
Thus, we use a substitution variable $z_{i,j}$ and an additional constraint $z_{i,j} \Leftrightarrow x_i \land y_j$ for every pair $(x_i, y_j)$.
The added constraint ensures that $z_{i,j}$ behaves equivalently to $x_i \cdot y_i$.
For a \emph{division} of two sums, our encoding depends on enumerating the variable selections of the divisor.
The main problem lies in the following partial construct after resolving the division: $\frac{k_i \cdot x_i}{c_2 + \sum^m_{i=1} l_i \cdot y_i}$.
To get the required structure, we create a sum with each element representing the fraction under one variable selection $s \in S$: $\sum_{s \in S} \frac{k_i}{c_2 + \sum_{i \in s} l_i} \cdot e_s$.
Each substitution variable $e_s$ is equivalent to the corresponding selection $s$ (\ie, $e_s \Leftrightarrow x_i \wedge \bigwedge_{i \in s} y_i \bigwedge_{\bar{i} \in s} \neg y_i$).
In addition to that, we create pseudo-Boolean constraints that mark each assignment causing division by zero as unsatisfied (\ie, $s \Leftrightarrow \sum_{i=1}^n l_i \cdot y_i \neq -c_2$).
Both our encodings for multiplication and division require the introduction of artificial variables, but the encodings preserve quasi-equivalence~\cite{KKS+:ASE22}.
Hence, when compiling the resulting formula to d-DNNF later, the d-DNNF produces consistent results.

\paragraph{Normalizing Pseudo-Boolean Constraints}
Some provided encodings slightly deviate from the structure of pseudo-Boolean constraints presented in \autoref{subsec:background:pseudoboolean}.
The definition enforces $\geq$ as comparison operator, but our encodings also use $=$ and attribute constraints in UVL supports all the common comparison operators (\ie, =, $\geq$, $\leq$, $<$, $>$, and $\neq$).
For all operators but $\neq$, we use the conversions shown in \autoref{table:normalization} to conform with the definition.
In contrast, we keep $\neq$ constraints and handle them in our compiler as adapting them to match the structure in \autoref{equation:pseudoboolean} is complex.
Furthermore, we convert floats to integers by multiplying both sides of the inequality with $10^X$ where $x$ is the highest number of decimal places~\cite{CK:CADICS05}.

\begin{table}[htbp]
  \centering
  \caption{Converting Comparison Operators \todo{cite}}
  \label{table:normalization}
  \begin{tabular}{ll}
    \toprule
    Origin                              & Normalized                                                                    \\
    \midrule
    $\sum_{i=1}^n a_i \cdot v_i \leq b$ & $\sum_{i=1}^n -a_i \cdot v_i \geq -b$                                         \\
    $\sum_{i=1}^n a_i \cdot v_i > b$    & $\sum_{i=1}^n a_i \cdot v_i \geq b + 1$                                       \\
    $\sum_{i=1}^n a_i \cdot v_i < b$    & $\sum_{i=1}^n -a_i \cdot v_i \geq -b + 1$                                     \\
    $\sum_{i=1}^n a_i \cdot v_i = b$    & $\sum_{i=1}^n -a_i \cdot v_i \geq -b \land \sum_{i=1}^n a_i \cdot v_i \geq b$ \\
    \bottomrule
  \end{tabular}
\end{table}

\paragraph{Summary}
By combining the different translation strategies, we can now encode all constructs from feature models with basic constructs $\mathbb{B}$, cardinality constructs $\mathbb{C}$, and attribute constraints $\mathbb{A}$ in pseudo-Boolean logic.
For various constructs, the proposed pseudo-Boolean encoding is more sparse than the respective Boolean encoding.
The differences are especially apparent for cardinalities and addition/subtraction based attribute constraints, where we achieve linear-sized encodings in pseudo-Boolean, but the Boolean encoding grows exponentially in size.
Note that we examine the correctness and efficiency in practice of our encodings empirically in \autoref{sec:evaluation}.
Having the encoding enables us to employ pseudo-Boolean based reasoning on UVL models, including our proposal; \textit{pseudo-Boolean d-DNNF compilation} (\cf \autoref{sec:compilation}).

\section{Pseudo-Boolean d-DNNF Compilation}\label{sec:compilation}
Our goal is to provide scalable reasoning for common feature-modeling constructs that are hard to analyze with state-of-the-art reasoning engines which typically use CNFs as input~\cite{B:JSAT08,LP:JSAT10,LGCR:SPLC15,MMC+:DAC01,KTS+:VaMoS20}.
To this end, we propose pseudo-Boolean d-DNNF compilation, which comes with two main advantages.
First, while the input is \emph{pseudo-Boolean}, our approach produces \emph{Boolean} d-DNNFs which can be used with available strategies that enable a plethora of feature-model analyses~\cite{SRH+:TOSEM24,BDD+:TR23,SGRM:LPAIR18}.
Second, pseudo-Boolean formulas can provide more sparse encodings for several constructs compared to CNF, while still only having to manage Boolean variables.
The limitation to Boolean variables enables us to employ popular techniques from Boolean d-DNNF compilation with only slight adaptations.
For instance, exhaustive DPLL~\cite{BL:JAIR99}, which is the algorithmic basis of every popular Boolean d-DNNF compiler~\cite{D:AAAI02,MMBH:AAI12,LM:IJCAI17}, is not applicable for non-Boolean variables.

\subsection{Reusing Ideas from Boolean Compilation}
We propose pseudo-Boolean compilation based on exhaustive DPLL~\cite{BL:JAIR99} which is also the de facto standard for Boolean compilation (\cf \autoref{subsec:background:ddnnfcompile})~\cite{LM:IJCAI17,MMBH:AAI12,D:AAAI02}.
On the abstraction level of \autoref{algorithm:booleanddnnf}, the algorithms for pseudo-Boolean and Boolean (\ie, using CNFs as input) are equivalent.
However, many details need to be considered and changed for pseudo-Boolean d-DNNF compilation.

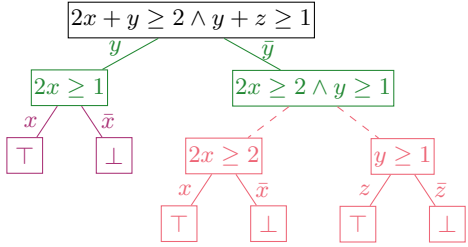
\begin{figure}[t]
  \centering
  \begin{forest}
    for tree={
    font = \small,
    grow=south,
    rectangle, draw, minimum size=3ex, inner sep=1pt,
    s sep=7mm
    }
    [$2x + y \geq 2 \land y + z \geq 1$, for children={ tolbrightgreen, edge=tolbrightgreen}
      [$2x \geq 1$, for children={ tolbrightpurple, edge=tolbrightpurple}, edge label={node[midway,above,xshift=-0.2cm,yshift=-0.15cm, font=\small]{$y$}}
          [$\top$, edge label={node[midway,left,font=\small]{$x$}}]
          [$\bot$, for children={ tolbrightblue, edge=tolbrightblue}, edge label={node[midway,right,font=\small]{$\bar{x}$}}]
      ]
      [$2x \geq 2 \land y \geq 1$,for children={ tolbrightred, edge={dashed,tolbrightred}}, edge label={node[midway,right, yshift=0.05cm,font=\small]{$\bar{y}$}}
          [$2x \geq 2$, for children={ tolbrightred, edge=tolbrightred}
              [$\top$, edge label={node[midway,left,font=\small]{$x$}}]
              [$\bot$, edge label={node[midway,right,font=\small]{$\bar{x}$}}]
          ]
          [$y \geq 1$, for children={ tolbrightred, edge=tolbrightred}
              [$\top$, edge label={node[midway,left,font=\small]{$z$}}]
              [$\bot$, edge label={node[midway,right,font=\small]{$\bar{z}$}}]
          ]
      ]
    ]
  \end{forest}
  \caption{Trace for Exhaustive Pseudo-Boolean DPLL}
  \label{figure:exhaustivepseudo}
\end{figure}

\autoref{figure:exhaustivepseudo} shows an example for our adaptation of exhaustive DPLL to a pseudo-Boolean formula.
The adaptation follows previous adaptations of \emph{non-exhaustive} DPLL for pseudo-Boolean formulas~\cite{CK:CADICS05}.
We recursively branch on the variables $x, y, z$ until we have a conflict (\ie, at least one constraint cannot be satisfied by extending the current assignment) or all constraints are satisfied.
Each time we assign a variable, we have to update every constraint that contains the assigned variable.
For instance, $2x + y \geq 2 \land y + z \geq 1$ becomes $2x \geq 1$ under the assignment $y = 1$.
When we reach a subformula, where we can split the constraints in a way that they share no variables, we traverse those splits separately as seen in \autoref{figure:exhaustivepseudo} for $2x \geq 2 \land z \geq 1$.

\begin{figure}
  \centering
  \begin{tikzpicture}[dag,font=\small]
    \node[tolbrightgreen] {$\lor$}
    child[tolbrightgreen] {
        node {$\land$}
        child {
            node {$y$}
          }
        child[tolbrightpurple] {
            node {$x$}
          }
      }
    child[tolbrightgreen] {
        node {$\land$}
        child {
            node {$\bar{y}$}
          }
        child[tolbrightred] {
            node {$\land$}
            child {
                node {$x$}
              }
            child {
                node {$z$}
              }
          }
      };
  \end{tikzpicture}
  \caption{d-DNNF Corresponding to Traversal in \autoref{figure:exhaustivepseudo}}
  \label{figure:pseudoddnnf}
\end{figure}
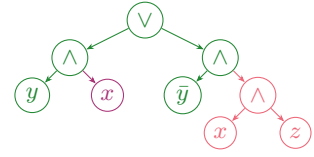

As the computed trace (\autoref{figure:exhaustivepseudo}) only contains Boolean decisions, we can, analogously to Boolean compilation, derive a d-DNNF from it. \autoref{figure:pseudoddnnf} shows the resulting d-DNNF.
So, we get semantically equivalent results to the CNF-based state-of-the-art~\cite{LM:IJCAI17}, but benefit from smaller input formulas for several feature-modeling constructs.
For every assignment, we create a disjunction, which is, by construction, deterministic.
Every time we split the formula due to disconnected components, we create a conjunction which is decomposable by construction.
Note that we already propagated the $\top$ and $\bot$ nodes from \autoref{figure:exhaustivepseudo}.
For instance, we simplified $(x \land \top) \lor (\bar{x} \land \bot)$ to $x$.

\subsection{Optimizations}
The scalability of DPLL-based reasoning engines heavily relies on optimizations~\cite{T:SAT06,KJ:TR23,LM:IJCAI17,FHI+:AIJ21,FHH:JEA21}.
Over the last decades, various strategies for Boolean reasoning (\eg, SAT solving, model counting, and knowledge compilation) have been proposed.
In the following, we briefly introduce some popular strategies that we have adopted for our pseudo-Boolean approach.
Furthermore, we describe the required adaptations.

\paragraph{Boolean Constraint Propagation}
Boolean constraint propagation refers to automatically assigning variables for which the formula is only satisfiable under that specific assignment.
In Boolean logic, this is detected via unit clauses (\ie, clauses with exactly one literal).
To satisfy the clause, the corresponding variable has to be assigned such that the literal is satisfied.
For pseudo-Boolean logic, we use an adaptation suggested by Chai and Kuehlmann~\cite{CK:CADICS05} to detect implied literals.
Given $\sum^n_{i=1} a_i \cdot v_i \geq b$, it can be checked if the largest constant $a_{max}$ is required to satisfy the inequality by checking $\sum^n_{i=1} a_i \cdot v_i < b + a_{max}$.
In this case, $v_{max}$ is implied.


\paragraph{Component Caching}
During DPLL traversal, the algorithm often produces the same component (\ie, subformula) multiple times.
With component caching, the goal is to save results of components to reuse them later~\cite{T:SAT06,SBB+:SAT04}.
The main problem here is to efficiently store and detect components, which is typically realized via hashing.
Popular hashing schemes used in Boolean solvers are not directly applicable for pseudo-Boolean formulas, because they rely on the structure of CNFs.
However, hashing the entire data structure of a subformula has a substantial impact on the performance.
We adapt the scheme used by sharpSAT~\cite{T:SAT06} to also include the right side of the pseudo-Boolean inequalities.
In particular, we encode a component as $((v_1,..v_n)(l, r_l)(k, r_k))$ to prepare for hashing where $v_i$ are unassigned variables, $l$ and $k$ are indices of unsatisfied constraints, and $r_l$ and $r_k$ their respective righthand sides. The lists are consistently sorted over the traversal.
For instance, we would encode the subformula $2x \geq 2 \land y \geq 1$ resulting from setting $\bar{y}$ as $((x), (1,2), (2,1))$, where the left side of the second and third tuple refers to the index of the original constraint.

\paragraph{Conflict Analysis}
In conflict analysis, the partial assignment leading to a conflict is saved as a conflict clause to prevent running into the same conflict multiple times~\cite{SBB+:SAT04}.
For instance, after finding that the assignment $\{y, \bar{x}\}$ is unsatisfiable, we could add its negation as conflict clause (\ie, $\neg(y \land \bar{x}) \equiv (\bar{y} \lor x)$) to detect this conflict earlier during further branching.
Common techniques aim to identify conflict clauses with as few as literals as possible as those can be potentially applied more often~\cite{MLM09, SBB+:SAT04}.
Since assignments are still Boolean in pseudo-Boolean logic and clauses can be easily encoded as pseudo-Boolean constraints, we can use Boolean conflict-driven clause learning (CDCL)~\cite{SBB+:SAT04,MLM09} with only minor technical adaptations.

\paragraph{Variable Orderings}
The ordering of variables during branching can have substantial impact on the compilation performance (\eg, regarding runtime)~\cite{LM:IJCAI17,D:AAAI02}.
Various strategies have been proposed that typically rely on the occurrence of variables in regular clauses (\eg, DLCS~\cite{SBK:SAT05}) and conflict clauses (\eg, VSIDS~\cite{MMC+:DAC01}).
For our compiler, we use the same primary variable sorting as \dfour{}~\cite{LM:IJCAI17}: prioritizing variables that result in disconnected components.
In previous work, we observed that \dfour{} performs best for compiling feature models~\cite{SKH+:AMAI24,SHN+:EMSE23}.
Analogous to \dfour{}, we translate our formula to a hypergraph and use an existing partitioner to derive a cut set, which we then use for ordering.
Within the cut set, we order the variables using an adaptation of VSIDS~\cite{MMC+:DAC01} which is commonly used in SAT and \ssat{}.
Our adapted version of VSIDS also considers the impact of a variable in a pseudo-Boolean constraint considering the degree of the formula and the factor of the variable (\cf \autoref{subsec:background:pseudoboolean}).
Furthermore, following insights from preliminary experiments, we use no no decay which would decrease the score of variables that are not part of conflict clauses over time~\cite{MMC+:DAC01}.

\subsection{Tooling}
We provide publicly available tooling for both major steps of our pipeline, namely encoding feature models in pseudo-Boolean logic and pseudo-Boolean d-DNNF compilation.
The encoder \uvltopb{}\footnote{\url{https://github.com/TUBS-ISF/pseudo-boolean-uvl-encoder}} is implemented in Java, so it can be directly used with the UVL parser.\footnote{\url{https://github.com/Universal-Variability-Language/uvl-parser}}
The output \texttt{.opb} (standard format for pseudo-Boolean formulas) file can then be used as input for our compiler \ptod{}\footnote{\url{https://github.com/TUBS-ISF/p2d}}, whose name is a tribute to the d-DNNF compiler \ctod{} that compiles CNFs to d-DNNF.
The pseudo-Boolean compiler \ptod{} can be used for d-DNNF compilation and for model counting.
The compiled d-DNNFs can be saved in the format proposed by \dfour{}~\cite{LM:IJCAI17}, which can then be used as input for further reasoning (\eg, \ddknnife{}~\cite{SRH+:TOSEM24} or \texttt{Winston}~\cite{BDD+:TR23}).

\section{Evaluation} \label{sec:evaluation}

With our empirical evaluation, we examine the benefits of our approach compared to the state-of-the-art.
To this end, we compare the performance of pseudo-Boolean with Boolean d-DNNF compilation, including our pseudo-Boolean encoding of the feature model.
For the performance, we consider representation size and runtime for both encoding and d-DNNF compilation.
We provide a reproduction package including the evaluated tools, subject systems, and final results.\footnote{\url{https://github.com/TUBS-ISF/pb-ddnnf-eval} \textcolor{blue}{(Zenodo for final version)}}

\subsection{Experiment Design}

\paragraph{Research Questions}
To examine the advantages of our approach over the state-of-the-art, we evaluate the performance of the \emph{encoding} (\textbf{RQ1}) and the performance of the \emph{compilation} (\textbf{RQ2}, \textbf{RQ3}).

\newcommand{\rqone}{How does the performance of pseudo-Boolean encoding compare to CNF encoding for feature models?}
\newcommand{\rqtwo}{How does the performance of pseudo-Boolean d-DNNF compilation compare to Boolean compilation for industrial basic feature models?}
\newcommand{\rqthree}{How does the performance of pseudo-Boolean d-DNNF compilation compare to Boolean compilation for expressive constructs?}

\begin{description}[leftmargin=\researchquestionlabelwidthof{1}]
  \researchquestion{1} \rqone{}
\end{description}

\begin{description}[leftmargin=\researchquestionlabelwidthof{2}]
  \researchquestion{2} \rqtwo{}
\end{description}

\begin{description}[leftmargin=\researchquestionlabelwidthof{3}]
  \researchquestion{3} \rqthree{}
\end{description}

For the performance, we consider two dimensions: size of representation \wrt to number of literals and runtime for compiling/encoding the input.
With \textbf{RQ1}, we compare the runtimes required to encode the feature models in Boolean and pseudo-Boolean and the sizes of the resulting formulas.
In \textbf{RQ2}, we examine the performance of our approach on feature models with only basic constructs for two reasons.
First, while those constructs are used in practice~\cite{RFB+:MODEVAR22,DGR:AUSE11,gears,HBH:SLE11,PNX+:FOSD11, BNR+:MODELS14, WAYL:QSIC13, BTS:SEFM19}, there are currently no industrial feature models with expressive constructs \emph{publicly} available.
Second, as basic feature-modeling constructs are widely used across systems and application domains~\cite{SBK+:SPLC24,LSB+:SPLC10,HST+:MODELS22,BSL+:TSE13,MBC:OOPSLA09}, we expect that many feature models with more expressive constructs still contain a considerable share of basic constructs.
Hence, the performance on basic models is still relevant for feature models containing expressive constructs.
Even though there is a lack of publicly available feature models with the expressive constructs, we aim to provide indicative insights on how our approach performs for expressive constructs.
With \textbf{RQ3}, we compare the performance of the pseudo-Boolean and Boolean approach for compiling feature models representing various combinations of constructs.

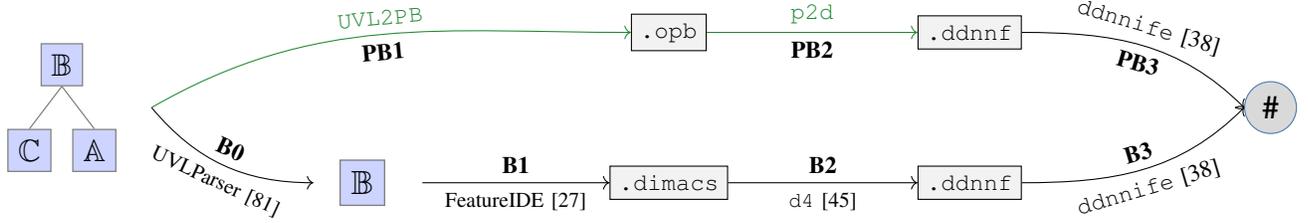
\begin{figure*}
  \centering
  \normalsize
  \begin{tikzpicture}[font=\small]
    \node (FM) at (0,0) {\includegraphics[width=2cm]{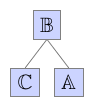}};

    \node (cFM) at (4,-1) {\includegraphics[width=1.2cm]{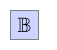}};
    \node[draw,fill=gray!10] (pb) at (8,1) {\texttt{.opb}};

    \draw[->,tolbrightgreen] (FM.east) to[in=180, out=30] node[midway, sloped, above] {\texttt{UVL2PB}} node[midway,sloped, below] {\textcolor{black}{\textbf{PB1}}} (pb.west);
    \draw[->] (FM.east) to[in=180,out=310] node[midway, sloped, below] {\footnotesize UVLParser~\cite{SVT+:SPLC23}} node[midway,sloped, above] {\textbf{B0}} (cFM.west);

    \node[draw,fill=gray!10] (CNF) at (8, -1) {\texttt{.dimacs}};

    \draw[->] (cFM.east) -- node[midway, sloped, below] {\footnotesize FeatureIDE~\cite{MTS+17}} node[midway,sloped, above] {\textbf{B1}} (CNF.west);

    \node[draw,fill=gray!10] (pddnnf) at (12, 1) {\texttt{.ddnnf}};
    \node[draw,fill=gray!10] (cddnnf) at (12, -1) {\texttt{.ddnnf}};

    \draw[->,tolbrightgreen] (pb.east) -- node[midway, sloped, above] {\texttt{p2d}} node[midway,sloped, below] {\textcolor{black}{\textbf{PB2}}} (pddnnf.west);
    \draw[->] (CNF.east) -- node[midway, sloped, below] {\footnotesize \dfour{}~\cite{LM:IJCAI17}}  node[midway,sloped, above] {\textbf{B2}}(cddnnf.west);

    \node[draw=tolbrightblue,circle,fill=gray!30] (ddnnife) at (16, 0) {\large \textbf{\#}};

    \draw[->] (pddnnf.east) to[out=0, in=135] node[midway, sloped, above] {\ddknnife{}~\cite{SRH+:TOSEM24}}  node[midway,sloped, below] {\textbf{PB3}} (ddnnife.west);
    \draw[->] (cddnnf.east) to[out=0, in=225] node[midway, sloped, below] {\ddknnife{}~\cite{SRH+:TOSEM24}}  node[midway,sloped, above] {\textbf{B3}} (ddnnife.west);

  \end{tikzpicture}
  \caption{Comparison of Boolean and Pseudo-Boolean Approach for Compiling Expressive Feature Models to d-DNNF}
  \label{figure:evaluation:architecture}
\end{figure*}

\paragraph{Evaluated Tools}
With the evaluation, we aim to compare our approach to the Boolean state-of-the-art which is CNF to d-DNNF compilation~\cite{SKH+:AMAI24,LM:IJCAI17}.
\autoref{figure:evaluation:architecture} provides an overview on the pseudo-Boolean and Boolean pipeline.
Our contributions are indicated with \textcolor{tolbrightgreen}{green} arrows.
For our approach, we evaluate our new proposals, namely \uvltopb{} for the pseudo-Boolean encoding and \ptod{} for the d-DNNF compilation.
For the Boolean baseline, we need a way to translate the models with expressive constructs to CNF.
To this end, we use UVL conversion strategies~\cite{SVT+:SPLC23} to translate the expressive models to basic models.
Then, we use FeatureIDE's CNF translation, as it supports conversion from UVL and is commonly employed in other evaluations~\cite{SRH+:TOSEM24, BDD+:TR23,KAT:GPCE16,KKS+:ASE22,FSRR:VaMoS21}.
For Boolean d-DNNF compilation, we use \dfour{}\footnote{\url{https://github.com/crillab/d4v2}} as other evaluations clearly indicate that \dfour{} performs best for feature-model analyis~\cite{SRH+:TOSEM24,SHN+:EMSE23}.

\paragraph{Subject Systems}
Currently, there is a lack of industrial feature models with expressive constructs in the literature.
Even though commercial tools support expressive constructs~\cite{RFB+:MODEVAR22,DGR:AUSE11,gears} and such constructs are used in practice~\cite{HBH:SLE11,WAYL:QSIC13,BNR+:MODELS14}, feature models created with these tools are not \emph{publicly} available.
With our selection of subject systems, we aim to provide indicators on the performance of our approach in practice by evaluating on real-world basic feature models and synthesized feature models with expressive constructs.
\autoref{table:evaluation:subjectsystems} gives an overview on the different datasets.
Each evaluated feature model is given in UVL format.
\todo{ek: what about Paul's courses~\cite{BTS:SEFM19}? cs: I think its fine that those are missing for now since they are no industrial configurable systems in that sense, but I think we should still add them to the evaluation for the next revision}

\begin{table}[htbp]
  \centering
  \caption{Overview Subject Systems}
  \label{table:evaluation:subjectsystems}
  \footnotesize
  \addtolength{\tabcolsep}{-0.2em}
  \begin{tabular}{lrrr}
    \toprule
    Dataset               & \texttt{\#}Models & \texttt{\#}Variables & Constructs                                              \\
    \midrule
    Literature Collection & 76                & 100--80,258          & $\mathbb{B}$                                            \\
    Confidential Industry & 2                 & 612--645             & $\mathbb{B}$ \& $\mathbb{A}$                            \\
    Single Construct      & 16,164            & 1--8,000             & Isolated from $\mathbb{C}$ \& $\mathbb{A}$              \\
    Random                & 2,700             & 80--2,767            & Combs. from $\mathbb{B}$, $\mathbb{C}$, \& $\mathbb{A}$ \\
    \bottomrule
  \end{tabular}
\end{table}

\noindent\emph{Real-World Basic Feature Models}
Most publicly available feature models are limited to basic constructs~\cite{SBK+:SPLC24,KTM+:ESECFSE17,OGB+:TR19,BSL+:TSE13}.
In addition, we expect that even feature models with more extended expressions contain a considerable share of basic constraints.
Hence, we evaluate our approach on real-world basic feature models.
We use a dataset recently collected from industrial models in the literature~\cite{SBK+:SPLC24}.
The full provided dataset contains several histories of feature models with hundreds of similar feature models.
For every history, we use three feature models with the minimum, median, and maximum number of constraints respectively instead of the entire history.
A configuration file specifying which models have been used can be found here.\footnote{\url{https://github.com/SoftVarE-Group/feature-model-benchmark/blob/master/pre_configs/minmedmax.json}}
The used dataset covers various domains (\eg, automotive and systems software) and varying sizes (100--80,258 variables and 0--388,816 constraints).

\noindent\emph{Industry Models with Attribute Constraints.}
Unfortunately, there is a lack of real-world feature models with expressive constructs in the literature.
From our industry partner, we have access to two feature models that each contain numerous numeric constraints over feature attributes. Overall, the feature models include constructs from basic feature models $\mathbb{B}$ and attribute constraints $\mathbb{A}$, but no cardinalities ($\mathbb{C}$).

\noindent\emph{Single Construct Models.}
With the third dataset, we examine the performance of our pseudo-Boolean pipeline for different language constructs in isolation.
In particular, we generate feature models including only one of alternative, feature cardinality, group cardinality, and numeric constraints and if needed optional features.
\autoref{table:eval:constructmodels} provides an overview on the generated models for the specific constructs.
For each incarnation of a specific construct, we create a feature model only including features and constraints corresponding to this incarnation.
Depending on the construct, we create different combinations of relevant parameters to achieve a wide coverage.
As an example, for alternatives, we create 80 feature models consisting of one alternative with between 100 and 8,000 features each.
To evaluate feature cardinality, we create a single feature with that cardinality and attach four groups, each having one type of the \emph{basic} hierarchical groups and three child features.
Here, we vary the number of features in the group, the lower bound of the cardinality, and the upper bound.
For the numeric constraints, we evaluate constraints consisting of growing expressions $A$ of the respective operators and a constant $\mathbb{E}_p(A)$ for comparison.
The constant $\mathbb{E}_p(A)$ controls that a share $p$ of the features need to be selected on average to satisfy the constraint.
So for $p = 0.1$, the value $\mathbb{E}_p(A)$ is set to a value that is met or exceeded when selecting 10 \% of the features on average.

\noindent\emph{Generated UVL Models.}
To examine the performance of the different approaches, when dealing with different combinations of the constructs, we randomly generate further UVL models using the UVLGenerator~\cite{SHS+:MODEVAR24b}.\footnote{https://github.com/SoftVarE-Group/uvlgenerator}
To this end, We aim to cover a wide range of structural properties and distribution of constructs.
We generate feature models that include basic constructs and additional expressive constructs by following three strategies: (1) \emph{exactly one} expressive construct, (2) \emph{all but one} expressive construct, (3) all expressive constructs.
For each set of included expressive construct, we create 300 feature models which are separated in three size clusters: small, medium, and large.
The size clusters are derived from statistics over the collection of real-world feature models~\cite{SBK+:SPLC24}.
We use the minimal\footnote{In the collection~\cite{SBK+:SPLC24}, only feature models with 100 or more features are considered.} (100), median (554), and third quartile (2,306) number of features from the collection as point of reference for the small, medium, and large models.
The generated feature models have the respective reference value +/- 20~\% number of features.
We configured other structural properties (\eg, tree depth and distribution of group types) in the configurator to match the corresponding distributions of that property in the feature-model collection~\cite{SBK+:SPLC24}.

\begin{table}
  \newcommand{\ceil}[1]{\left\lceil #1 \right\rceil}
  \centering
  \caption{Overview Single Construct Models}
  \label{table:eval:constructmodels}
  \addtolength{\tabcolsep}{-0.2em}
  \footnotesize
  \begin{tabular}{lr}
    \toprule
    Construct           & How?                                                                                                            \\
    \midrule
    Alternative         & $\mathit{alt}^p_{c_1,..c_n}$ with $n \in \{100, 200, \ldots, 8\,000\}$                                          \\
    \midrule
    Group cardinality   & $[x..y]^p_{c_1,..,c_n}$ with $\forall x, y : x < y$                                                             \\
    Feature cardinality & $f[\ceil{x}..\ceil{y}]$ with $\forall x, y : x < y$                                                             \\
                        & $x, y \in \{1, 0.25n, 0.5n, 0.75n, n\}$                                                                         \\
                        & $n \in \{1,2,\ldots, 100, 200, \ldots 2\,000\}$
    \\
    \midrule
    Numeric constraints & $\sum_{i=1}^n a_i \geq b$ with $b = \mathbb{E}_p(\sum_{i=1}^n a_i)$                                             \\
                        & $n \in \{1,2,\ldots, 100, 200, \ldots 2\,000\}$                                                                 \\
                        & $\prod_{i=1}^n a_i + 1 \geq b$ with $b = \mathbb{E}_p(\prod_{i=1}^n a_i + 1)$                                   \\
                        & $n \in \{2,3,\ldots, 18\}$                                                                                      \\

                        & $\frac{a_1 \star \ldots \star a_{\nicefrac{n}{2}}}{a_{\nicefrac{n}{2} + 1} \star \ldots \star a_n} > b$ with $b
    = \mathbb{E}_p(\frac{a_1 \star \ldots \star a_{\nicefrac{n}{2}}}{a_{\nicefrac{n}{2} + 1} \star \ldots \star a_n})$                    \\
                        & $n \in \{4,6,\ldots, 18\}$, $p \in \{0.1,\ldots,0.9\}$                                                          \\
    \bottomrule
  \end{tabular}
\end{table}

\paragraph{Experiment Execution}
For each of the feature models, we evaluate the whole pseudo-Boolean and Boolean pipeline (\cf \autoref{figure:evaluation:architecture}).
The pseudo-Boolean pipeline consists of translating the UVL model to pseudo-Boolean formula with our encoding (referred to as PB1), compile the formula with \ptod{} to d-DNNF (PB2), and compute the model counts produced by the d-DNNF with \ddknnife{} (PB3).
For the Boolean pipeline, we convert the UVL model to a basic feature model using UVL conversion strategies~\cite{BSF+:JSS25} (B0), translate the basic feature model to CNF with FeatureIDE~\cite{MTS+17} (B1), compile the CNF to d-DNNF with \dfour{}~\cite{LM:IJCAI17} (B2), and check the model count produced by the d-DNNF with \ddknnife{}~\cite{SRH+:TOSEM24} (B3).
The timeouts are ten minutes for conversion to pseudo-Boolean/CNF and ten minutes for compilation.
We collect the runtimes for the pipelines, which consist of PB1 + PB2 and B0 + B1 + B2, respectively.
To inspect whether a runtime difference is statistically significant, we use a Wilxocon signed-ranked test~\cite{W08} and assume the significance level $\alpha = 0.05$.
Furthermore, we inspect the sizes of the formula encodings from PB1 and B1 and of the compiled d-DNNFs from PB2 and B2.
As indicator for the correctness of our approach, we compare the model counts produced by \ddknnife{} for every model (\ie, PB3 vs B3).

\subsection{Results}
In the following, we present and discuss the results of our empirical evaluation.
For every successfully analyzed feature model, the d-DNNFs compiled by \ptod{} and \dfour{} produce the same model count.
Since we evaluate on 18,942 models with combinations of constructs, this is a strong indicator that both our pseudo-Boolean encoding and \ptod{} are correct.

\paragraph{Encoding (RQ1)}

\autoref{table:eval:encoding} provides an overview on the performance of the Boolean and pseudo-Boolean pipelines.
For each dataset, we compare the share of models that have been successfully encoded/compiled by each strategy and the respective sizes (median over models successfully encoded/compiled by both) of the input formula in CNF/pseudo-Boolean and the d-DNNFs.
The pseudo-Boolean encoding is significantly faster for every dataset, including the basic feature models.
The difference in encoding performance is especially apparent for group cardinality, where all isolated models with up-to 5,000 features could be encoded with pseudo-Boolean, but only 7.05 \% (up-to 13 features) with the Boolean encoding.
Similarly, all sums (up-to 2,100 features) could be encoded with pseudo-Boolean, but only 9.15\% (up-to 19 features) for Boolean.
For multiplication (94.7\% vs 60.9 \%) and division (100 \% vs 48.1 \%), the differences are smaller, but still considerably advantageous for pseudo-Boolean.
For the two expressive feature models from our industry partner, the pseudo-Boolean encoding required 1.44 and 1.51 seconds, while Boolean hit the timeout of ten minutes.
For alternatives, group cardinalities, and feature cardinalities, pseudo-Boolean produces over 90 \% fewer literals.
On the other side, for multiplication and division the Boolean encoding produces 98.4 \% and 83.6 \% smaller formulas.

\begin{table}
  \addtolength{\tabcolsep}{-0.3em}
  \footnotesize
  \caption{Boolean vs Pseudo Boolean: Performance Overview}
  \addtolength{\tabcolsep}{-0.135em}
  \label{table:eval:encoding}
  \begin{tabular}{lrrrrrrrr}
    \toprule
    \multirow{2}{*}{Dataset}       &
    \multicolumn{2}{c}{Enc. (\%)}  &
    \multicolumn{2}{c}{Enc. Size}  &
    \multicolumn{2}{c}{Comp. (\%)} &
    \multicolumn{2}{c}{d-DNNF Size}                                                                                                                        \\
                                   & CNF  & PB            & CNF         & PB             & \dfour{}      & \ptod{}       & \dfour{}       & \ptod{}        \\\midrule
    \textbf{Real-World Basic}      & 97.4 & 97.4          & 5,073       & \textbf{3.987} & \textbf{81.6} & 80.3          & \textbf{2,310} & 2,892          \\
    \midrule
    \multicolumn{9}{l}{\textbf{Isolated}}                                                                                                                  \\
    Alternative                    & 98.8 & \textbf{100}  & 1.6M        & \textbf{4,002} & 12.5          & \textbf{53.8} & 2,751          & 2,751          \\
    Group Cardinality              & 7.05 & \textbf{100}  & 42          & \textbf{22}    & 7.05          & \textbf{87.4} & 44             & 44             \\
    Feature Cardinality            & 79.1 & \textbf{99.0} & 42K         & \textbf{2,486} & 70.7          & \textbf{76.8} & 26k            & \textbf{1,800} \\
    Addition                       & 9.15 & \textbf{100}  & 37          & \textbf{26}    & 9.15          & \textbf{90.5} & \textbf{40.5}  & 59.5           \\
    Multiplication                 & 60.9 & \textbf{94.7} & \textbf{32} & 986            & 60.9          & \textbf{76.2} & \textbf{39}    & 308            \\
    Division                       & 48.1 & \textbf{100}  & \textbf{22} & 135            & 48.1          & \textbf{100}  & \textbf{20}    & 46             \\
    \midrule
    \multicolumn{9}{l}{\textbf{Synthesized}}                                                                                                               \\
    Just Group Card.               & 81.7 & \textbf{100}  & 9,709       & \textbf{8,667} & 81.7          & \textbf{100}  & 614            & \textbf{613}   \\
    Just Feature Card.             & 97.7 & \textbf{100}  & 41K         & \textbf{21K}   & 96.7          & 96.7          & 1,257          & \textbf{970}   \\
    Just Expression                & 100  & 100           & 9,860       & \textbf{9,191} & 100           & 100           & \textbf{686}   & 795            \\
    Just Aggregate                 & 3.67 & \textbf{100}  & 9,411       & \textbf{7,979} & 3.67          & \textbf{100}  & 489            & 489            \\
    All but Group Card.            & 2.0  & \textbf{100}  & 5,202       & \textbf{3,983} & 2.0           & \textbf{97.3} & 151            & \textbf{150}   \\
    All but Feature Card.          & 4.33 & \textbf{100}  & 2,140       & \textbf{1,964} & 4.33          & \textbf{99.3} & 125            & \textbf{113}   \\
    All but Expression             & 8.33 & \textbf{99.7} & 88K         & \textbf{35K}   & 8.33          & \textbf{98.3} & 1,313          & \textbf{872}   \\
    All but Aggregate              & 81.7 & \textbf{100}  & 34K         & \textbf{18K}   & 81.0          & \textbf{97.7} & 1,072          & \textbf{973}   \\
    All constructs                 & 6.0  & \textbf{100}  & 11K         & \textbf{5,384} & 6.0           & \textbf{96.3} & 276            & \textbf{186}   \\
    \midrule
    \textbf{Industrial}            & 0    & \textbf{100}  & N.A.        & \textbf{14K}   & 0             & 0             & N.A.           & N.A.           \\
    \bottomrule
  \end{tabular}
\end{table}

\researchanswer{\textbf{RQ1:} On every evaluated dataset, including basic feature models without any expressive constructs, pseudo-Boolean encoding is significanly faster than Boolean encoding. For some constructs, pseudo-Boolean encoding scales for thousands of features while Boolean encoding hits the timeout of ten minutes already for a few dozen features. The resulting pseudo-Boolean formulas are also smaller for every dataset except multiplications and divisions in isolation.}

\paragraph{Compiling Basic Feature Models (RQ2)}

\begin{figure}[htbp]
  \centering
  \includegraphics[width=\columnwidth]{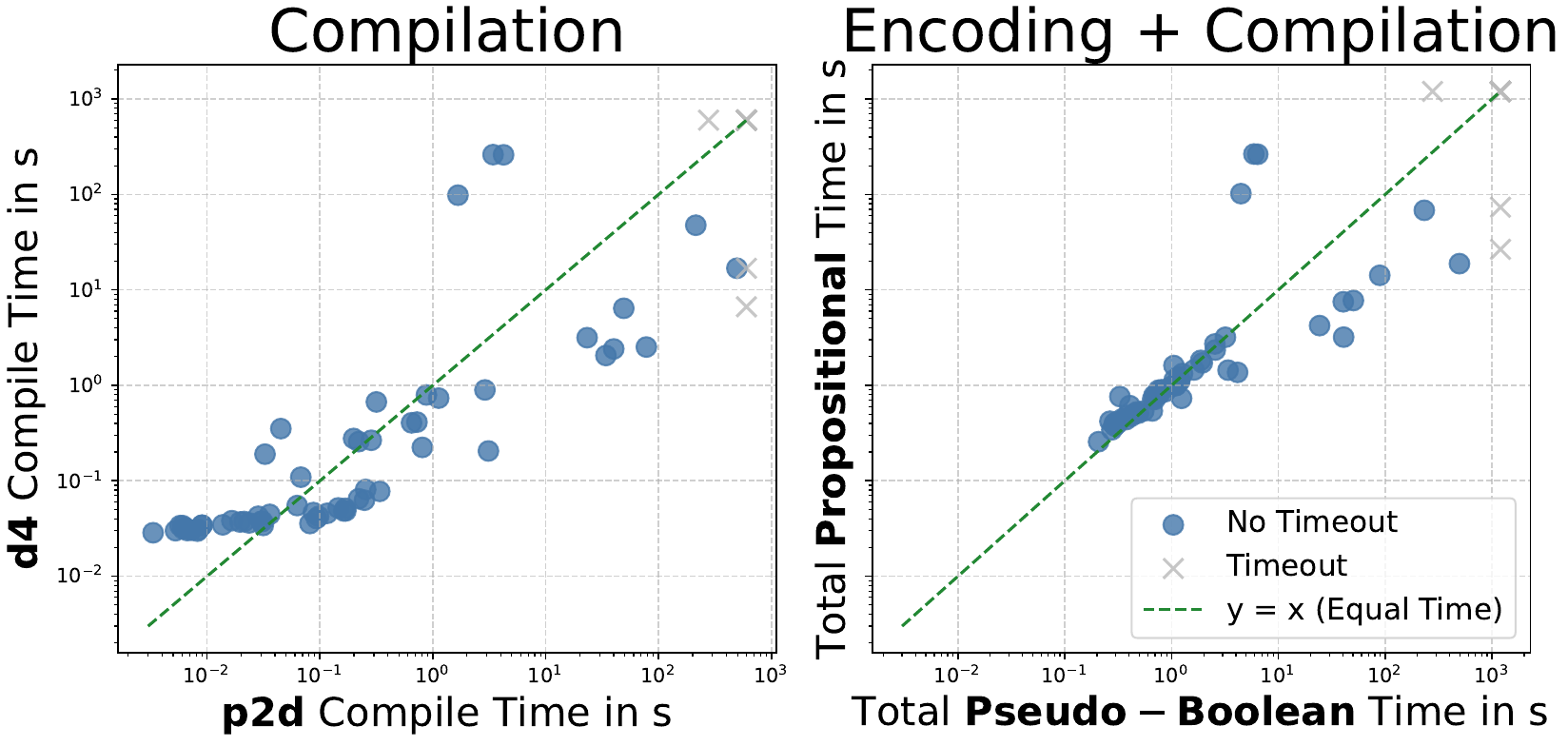}
  \caption{Compilation Comparison: Basic Feature Models}
  \label{figure:plot:basiccompilation}
\end{figure}

\autoref{figure:plot:basiccompilation} shows the runtimes of Boolean and pseudo-Boolean compilation without encoding time (left side) and with encoding time (right side).
Timeouts from at least one of the pipelines are marked with gray crosses \textcolor{tolbrightgray}{$\times$}.
For 11 out of the 74 successfully encoded feature models, both compilers resulted in a timeout.
Respectively, \ptod{} failed to compile for two other models and \dfour{} for one other.
\ptod{} was faster for 32 models and \dfour{} for 31.
\dfour{} required less overall runtime with 11.8 vs 15.9 minutes.
However, the difference between both compilers is not statistically significant ($p = 0.06$).
\dfour{} produces smaller d-DNNFs overall (2,310 vs 2,892 nodes in the median). \todo{ek: is this statistically significant?}

\researchanswer{\textbf{RQ2:} Our results suggest that our approach is competitive with state-of-the-art Boolean d-DNNF compilation even for basic feature models. Even though \dfour{} requires less runtime (but not significantly) for the dataset of basic feature models, \ptod{} is faster for about half of the models (32 of 63 successfully compiled).
  However, \ptod{} generally produces larger d-DNNFs, which may impact runtime of follow-up analyses.
  It may be valuable to optimize the size of compiled d-DNNFs in future work on \ptod{}.}

\paragraph{Compiling Expressive Constructs (RQ3)}

\begin{figure}[htbp]
  \centering
  \includegraphics[width=\columnwidth]{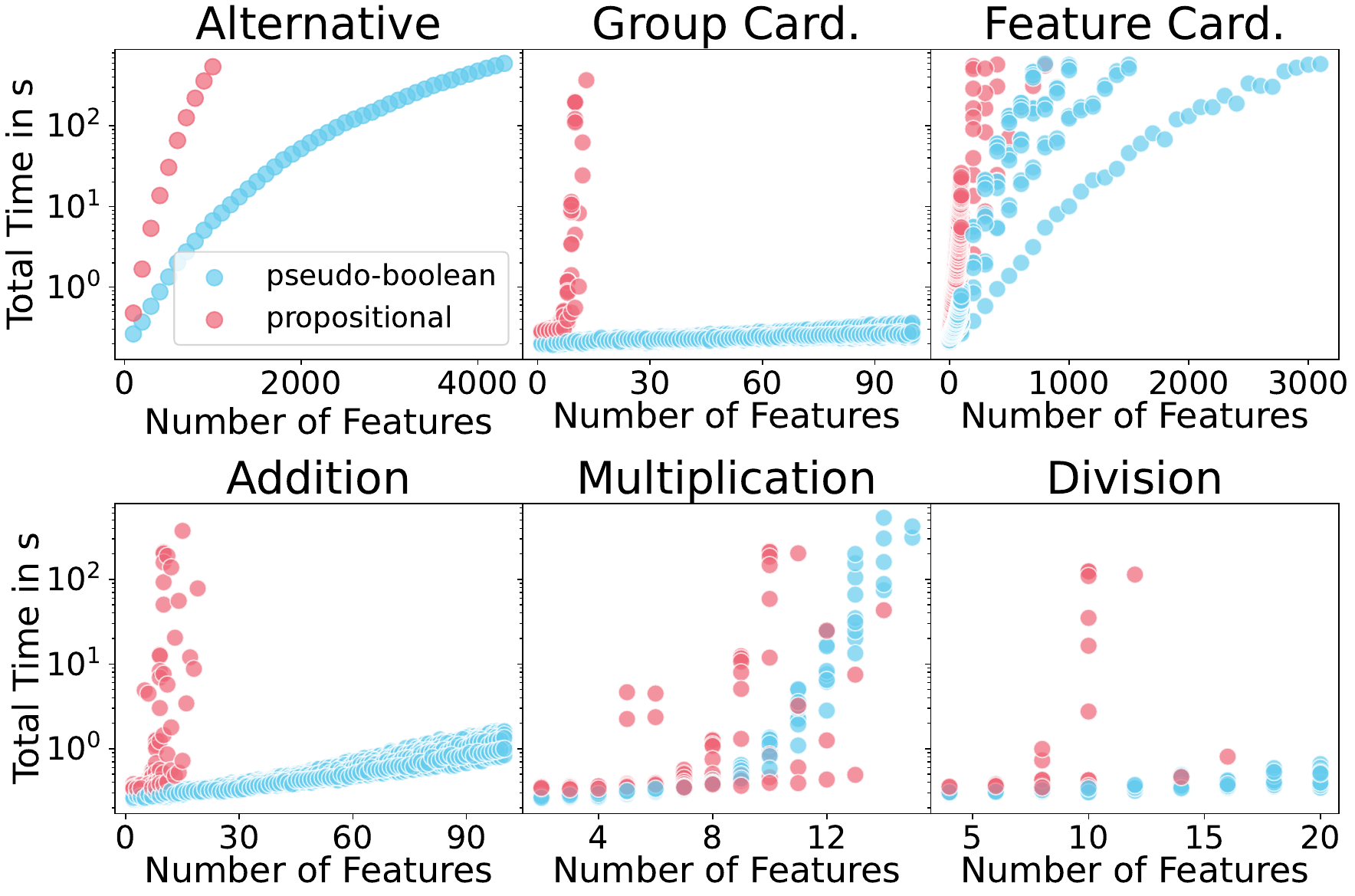}
  \caption{Compilation Comparison: Isolated Models}
  \label{figure:plot:isolated}
\end{figure}

\autoref{figure:plot:isolated} shows the runtimes of Boolean and pseudo-Boolean compilation including encoding times for the \emph{constructs in isolation}.
Here, the timeouts are omitted.
Note that for addition and group cardinality we only show an excerpt of the instances solved by \ptod{} as the performance of \dfour{} is hard to observe otherwise.
The pseudo-Boolean approach scales for a higher number of features for all constructs.
For group cardinality, \ptod{} successfully compiled 87.4 \% including ones from the largest category (5,000 features), while only 7.05 \% could be \emph{encoded in CNF}.
Nevertheless, each feature model encoded in CNF could be compiled to d-DNNF with \dfour{}.
For additions, the difference is also substantial as 90.5 \% (up-to 2,100 features) can be compiled with pseudo-Boolean, but only 9.15 \% (up-to 19 features) with the Boolean pipeline.
The differences for alternative (53.8 \% vs 12.5 \%), feature cardinality (76.8 \% vs 70.7 \%), multiplication (76.2 \% vs 60.9 \%) and division (100 \% vs 48.1 \%) are small but also considerable.

\begin{figure}[htbp]
  \centering
  \includegraphics[width=\columnwidth]{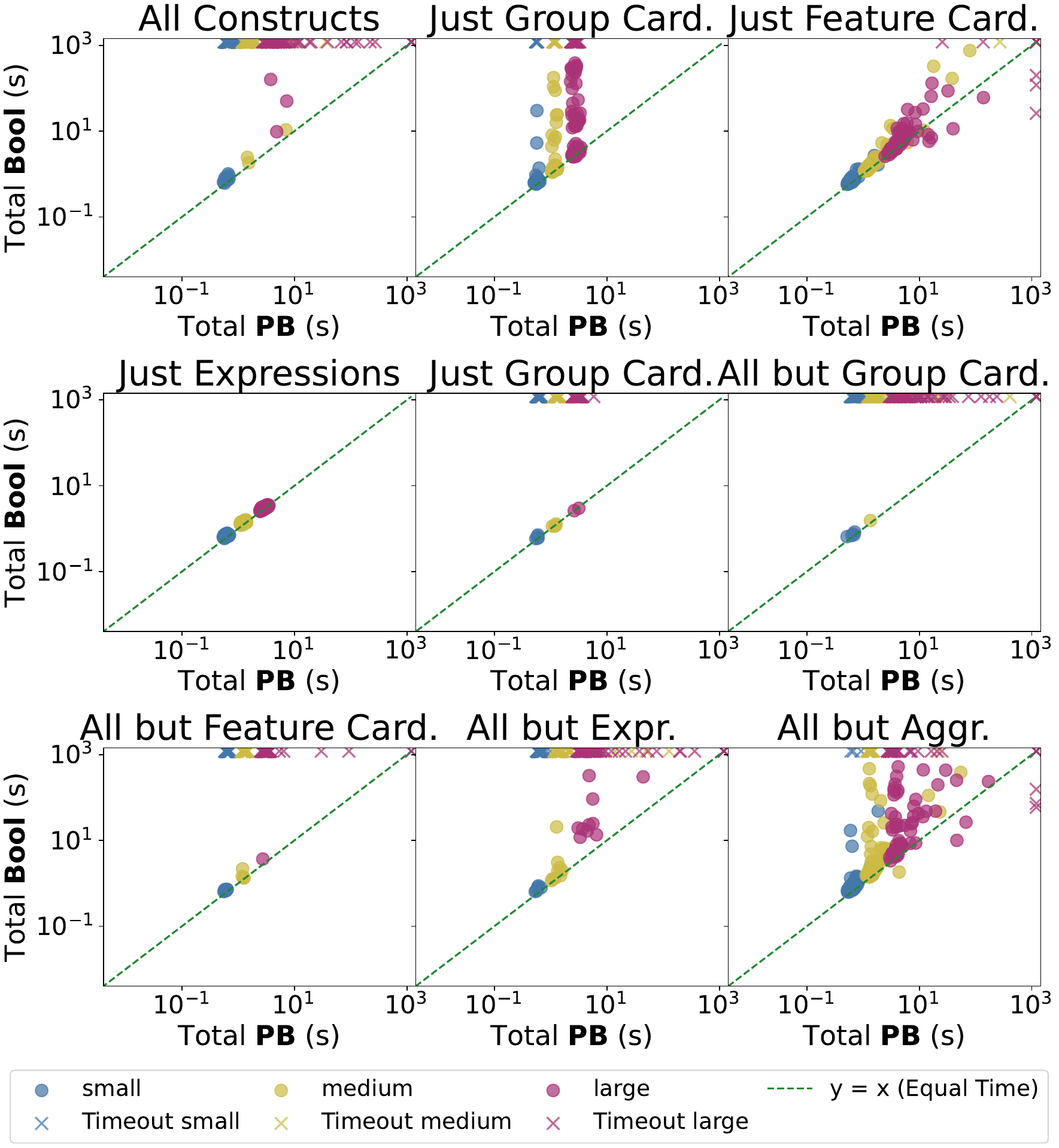}
  \caption{Compilation Comparison: Synthesized Models}
  \label{figure:plot:random}
\end{figure}

\autoref{figure:plot:random} shows the comparison for the different synthesized feature-model datasets.
\ptod{} is significantly faster for every dataset except \emph{just expressions} (\ie, feature model with basic constructs and numeric expressions over attributes) where it still requires less runtime overall.
Over the 2,700 analyzed feature models, the pseudo-Boolean pipeline successfully evaluated 
2,657 (98.4 \%) and the Boolean pipeline 1,156 (42.6 \%).
For both real-world models from our industry partner, compilation with \ptod{} hit the timeout of ten minutes, while for Boolean, already the encoding step failed.

The sizes of compiled d-DNNFs are generally similar, with small advantages for \ptod{} on most datasets.
However, for multiplication and division the d-DNNFs compiled by \ptod{} are considerably larger, probably due to the high number of artificial variables introduced by the encoding.
This is especially observable in the isolated models where the d-DNNFs compile in the median to 308 (\ptod{}) vs 39 (\dfour{}) nodes for multiplication and 46 vs 20 nodes for division.

\researchanswer{\textbf{RQ3:} For expressive feature-modeling constructs, pseudo-Boolean d-DNNF compilation yields substantial runtime benefits over the Boolean state-of-the-art. Especially if feature models contain aggregate functions or group cardinalities, our pseudo-Boolean approach substantially outperforms the Boolean baseline. Still, the results on the two industrial models and the encoding size for multiplication and division suggest that further optimization on our approach may be beneficial.}


\subsection{Threats to Validity}

\paragraph{Computational Bias}
The performance of running one of the tools can differ for the same input instances due to computational bias.
To reduce such a bias in our conclusions, we repeated each measurement three times and used the median.
The impact is further reduced by the deterministic behavior of each tool, also indicated by the little variance within the three repetitions.
Paired with the high number of evaluated models per tool, we do not suppose a considerable impact of computational bias on our conclusions.

\paragraph{Conversion to Logic}
Alternative strategies to translate the feature models to Boolean and pseudo-Boolean logic may impact the performance~\cite{KKS+:ASE22}.
For Boolean logic, a plethora of different encodings for constructs such as at-least-k and at-most-k~\cite{BTS:SEFM19}, or numerical variables~\cite{MPFB:JSS23} have been considered.
However, their suitability (\wrt semantic equivalence) or performance is unknown for encoding feature models and d-DNNF compilation.
We discuss this further in \autoref{sec:relatedwork}.
For pseudo-Boolean logic, we are not aware of alternative strategies in the literature to translate the constructs, but there should be numerous semantically equivalent encodings for the constructs.
However, adding more encodings for Boolean or pseudo-Boolean would add another layer of complexity to the already complex evaluation.
While out of scope for this work, considering further strategies may be valuable in the future for both Boolean and pseudo-Boolean encoding.

\paragraph{Tool Parameters}
Both compilers, \dfour{}~\cite{LM:IJCAI17} and \ptod{}, can be parametrized, which can have an impact on the performance.
However, considering the combinatorics of different parameters would also vastly increase the complexity of the evaluation and, thus, we consider it out of scope.
For the evaluation, we used the default parameters for every considered tool.

\paragraph{Transferability of Industrial Models}
Our selection of real-world models may not be transferrable to other feature models.
However, the collection used is the result of a broad literature survey to identify available real-world models in the literature~\cite{SBK+:SPLC24}.
The set covers various domains and properties, such as number of features, number of constraints, or tree structure.

\paragraph{Transferability of Synthesized Models}
The performance results on the synthesized models may not be transferable to practice.
However, we are not aware of industrial feature models with the expressive constructs available in the literature.
With our synthesized dataset, we aim to showcase the performance for different instances in isolation and also in combination with basic and other expressive constructs.
Furthermore, for the synthesized feature models, we considered structural properties of industrial feature models for the generation.
Nevertheless, we still consider an evaluation of our approach on more real-world models with expressive constructs as valuable future work.

\section{Related Work} \label{sec:relatedwork}

\paragraph{Pseudo-Boolean Solving}
Some pseudo-Boolean solvers have been proposed~\cite{CK:CADICS05,HK:JSAT06}.
However, these are limited to single-invocation satisfiability checks.
Very recently another pseudo-Boolean knowledge compiler has been proposed by Yang and Meel~\cite{YM:AAAI24}.
Their approach compiles to algebraic decision diagrams which are not limited to Boolean variables and generally more complex to compile.
In preliminary experiments, we compared their compiler to \ptod{}, but their compiler often produced inconsistent (compared to the results of \ptod{} and \dfour{}) model counts and is likely unsound.
Hence, we did not consider their compiler in our empirical evaluation.
Nevertheless, the preliminary experiments also showed that \ptod{} requires substantially less runtime for compiling the feature models.

\paragraph{d-DNNFs in Feature-Model Analysis}
d-DNNFs enable linear time (\wrt number of nodes in the d-DNNF) queries for satisfiability, counting, and enumeration~\cite{DM:JAIR02,SKH+:AMAI24}.
A plethora of feature-model analyses rely on potentially numerous of these operations~\cite{SKH+:AMAI24} and, thus, can benefit from d-DNNFs.
In related literature d-DNNFs have already been employed for counting~\cite{SRH+:TOSEM24,BDD+:TR23}, sampling~\cite{SGRM:LPAIR18,BDD+:TR23}, satisfiability checks~\cite{BDD+:TR23}, and deriving configurations optimized for different objectives~\cite{BDD+:TR23}.

\paragraph{Boolean d-DNNF Compilation}
Three Boolean d-DNNF compilers are commonly considered in the literature, namely \ctod{}~\cite{D:AAAI02}, \dsharp{}~\cite{MMBH:AAI12}, and \dfour{}~\cite{LM:IJCAI17}.
For our empirical evaluation, we only considered \dfour{} as baseline because the compiler substantially outperformed \ctod{} and \dsharp{} in previous evaluations on feature models~\cite{SKH+:AMAI24,SHN+:EMSE23}.
Our compiler \ptod{} adopts many of the strategies employed in the three Boolean compilers.
Still, the type of input formula is vital for the realization of a compiler, as internal datastructures and optimizations heavily rely on the exact structure (\ie, CNF for Boolean compilers).
Thus, it is not trivial to adapt an existing Boolean d-DNNF compiler to consider pseudo-Boolean formulas.

\paragraph{Expressive Feature-Model Dialects}
While reasoning for feature-model analysis mostly focuses on Boolean logic in CNF, more expressive constructs (\eg, group cardinality) were considered in a large variety of work. ter Beek~\etal~\cite{BSE:SPLC19} provide an overview on textual formats for specifying feature models. 11 out of 13 identified formats support at least one of the expressive constructs facilitated by pseudo-Boolean encoding.
Horcas~\etal~\cite{HPF:JSS23} suggest a metamodel approach to cover constructs from different variability languages. Their proposal also includes various constructs, such as arithmetic expressions and cardinalities.
In the initial publication of UVL, results of a community questionnaire imply that expressive constructs such as cardinality groups and feature attributes are desired~\cite{SFE+:SPLC21}.
Various variability-modeling tools that are commonly used in practice~\cite{BRN+:VaMoS13}, such as \texttt{pure::variants}~\cite{RFB+:MODEVAR22}, DOPLER~\cite{DGR:AUSE11}, and Gears~\cite{gears} also supports more expressive constraints.
Furthermore, several case studies on industrial systems imply that the expressive constructs we targeted are mandated~\cite{HBH:SLE11,PNX+:FOSD11, BNR+:MODELS14, WAYL:QSIC13, BTS:SEFM19}.
In summary, insights from related work suggest that more expressive constructs are of interest for practitioners and researchers.
The plethora of variability languages, including UVL, can benefit from our advances in scalability for feature-model analysis on these expressive constructs.

\paragraph{Encodings for Expressive Feature-Modeling Constructs}
While knowledge compilation for expressive feature-modeling constructs has not been yet employed, there has been some works on \textit{encoding} these constructs.
Various publications present for cardinality constraints~\cite{BTS:SEFM19,BB:CP03,S:CP05} which can be used to model cardinalities.
However, many of these encodings only preserve equisatisfiability~\cite{BTS:SEFM19,BB:CP03,S:CP05}, which can produce faulty results for model counting and for subsequent analyses on a compiled d-DNNF.
Furthermore, the performance of these encodings when applied to feature models in general and as input for knowledge compilation is unknown.
Nevertheless, examining the plethora of available encodings may reveal suitable and possibly more efficient candidates.
We consider the analysis of existing encodings as valuable future work, but as out of scope for this work as it substantially increases the complexity of our evaluation.
Benavides~\etal~\cite{BTR:SEKE05} propose to encode extended (\ie, attributed) feature models with constraint programming. However, their approach only scales to very small feature models (\ie, fewer than 25 features), while our strategy allows analyzing feature models with thousands of features.
Karataş~\etal~\cite{KOD:SCP13} also encode extended feature models with cardinality groups with constraint programming.
However, their approach also employs one-shot computations instead of knowledge compilation.
Munoz~\etal~\cite{MPFB:JSS23} employ bit-blasting to represent numeric features and constraints over them in Boolean logic.
However, they did not consider other expressive constructs such as cardinalities.
While we did not consider numeric features in this work, the bit-blasting strategy could also be valuable to apply for pseudo-Boolean logic.

\section{Conclusion}
There is a mismatch between the expressiveness of available variability languages~\cite{BSE:MODEVAR19} and available reasoning engines~\cite{SHN+:EMSE23,LGCR:SPLC15}. Most variability languages allow various expressions~\cite{BSE:MODEVAR19} that are not well suited for Boolean encoding. However, state-of-the-art feature-model analysis is mainly focused on Boolean logic~\cite{SKH+:AMAI24,MPFB:JSS23,LGCR:SPLC15,MTS+17,GHF+:SPLC23,ACLF:SCP13,MBC:OOPSLA09}.
Matching expectations and previous observations~\cite{BTS:SEFM19}, our evaluation provides more evidence that encoding common constructs, such as cardinalities or attribute constraints, in Boolean logic comes with substantial scalability issues.

In our work, we tackle this issue by introducing pseudo-Boolean d-DNNF compilation.
Our approach consists of a pseudo-Boolean encoding for feature models and a pseudo-Boolean d-DNNF compiler.
As the compiled d-DNNF is Boolean, we can employ existing algorithms and tools that enable efficient feature-model analyses~\cite{SRH+:TOSEM24,BDD+:TR23,SGRM:LPAIR18}.
Our empirical evaluation clearly suggests that using our pseudo-Boolean approach yields substantial runtime benefits over the Boolean state-of-the-art.
We did not tackle all feature-modeling constructs, as, for instance, we are not able to naturally represent features with other types (\eg, numeric) with pseudo-Boolean logic.
Still, pseudo-Boolean d-DNNF compilation considerably advances what feature-modeling constructs can be efficiently analyzed.
We envision that the availability of scalable reasoning increases the usage of these constructs, as automated analysis is vital for product-line engineering~\cite{BSRC10,TAK+:CSUR14,HFCA:IJSEKE13,KZK:LoCoCo10,HST+:MODELS22,VAT+:SPLC18,OPS+:REJ17}.


\bibliography{MYabrv, literature-cleaned, toadd}
\bibliographystyle{IEEEtran}

\newpage


\vspace{11pt}

\vfill

\end{document}